\documentclass[runningheads]{llncs}

\PassOptionsToPackage{table}{xcolor}
\usepackage[mobile]{eccv}

\usepackage{eccvabbrv}
\usepackage{siunitx}
\newcommand{\best}[1]{\textbf{#1}}

\usepackage{graphicx}
\usepackage{booktabs}
\usepackage{arydshln}

\usepackage[accsupp]{axessibility}  % Improves PDF readability for those with disabilities.

\usepackage[breaklinks,colorlinks,citecolor=eccvblue]{hyperref}

\usepackage{orcidlink}

\usepackage{tikz}
\usepackage{amsmath,amssymb}
\usepackage{booktabs}
\usepackage{array}
\usepackage{float}
\usepackage{microtype}
\usepackage{xcolor}

\newcommand{\tighttable}{%
  \setlength{\tabcolsep}{3pt}%
  \renewcommand{\arraystretch}{1.1}%
}

\newcommand{\boxer}{Boxer\xspace}

\newcommand{\grayrow}{\rowcolor{gray!15}}

\newcommand{\myparagraph}[1]{\paragraph{\textbf{#1}}}
\begin{document}
\raggedbottom

% ---------------------------------------------------------------
% \title{\boxer: Lifting 2D Bounding Boxes to 3D from Egocentric Capture}
\title{\boxer: Robust Lifting of Open-World 2D Bounding Boxes to 3D}

% \textit{todo} REVIEW: If the paper title is too long for the running head, you can set

% \textit{todo} FINAL: Replace with your author list.
\author{Daniel DeTone \and
Tianwei Shen \and
Fan Zhang \and
Lingni Ma \and
Julian Straub \and
Richard Newcombe \and
Jakob Engel
}

% \textit{todo} FINAL: Replace with an abbreviated list of authors.
\authorrunning{D. DeTone et al.}

% \textit{todo} FINAL: Replace with your institution list.
\institute{Meta Reality Labs Research}
% \email{ddetone@uni-heidelberg.de}}

\maketitle
\vspace{-2.0em}

\begin{figure}[H]
  \centering
  \begin{tikzpicture}
    \node(p0){\includegraphics[width=0.98\textwidth]{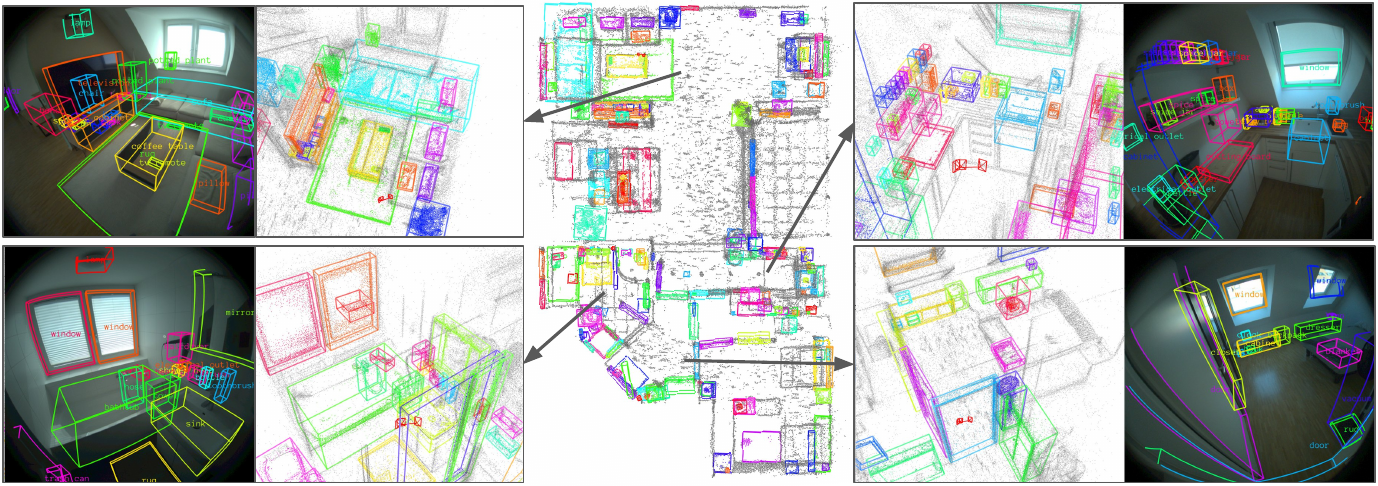}};
  \end{tikzpicture}
  \captionsetup{width=0.97\linewidth}
  \caption{Boxer takes as input posed images with optional depth and off-the-shelf 2DBB open-world detections, estimating static, global 3D bounding boxes. Boxer is run on a sequence and various scenes are highlighted to show the accuracy and open-world coverage of objects such as spice jar, hairdryer, sink drain and TV remote.}
  \label{fig:teaser}
\end{figure}

\begin{abstract}

Detecting and localizing objects in space is a fundamental computer vision problem. While much progress has been made to solve 2D object detection, 3D object localization is much less explored and far from solved, especially for open-world categories. To address this research challenge, we propose Boxer, an algorithm to estimate static 3D bounding boxes (3DBBs) from 2D open-vocabulary object detections, posed images and optional depth either represented as a sparse point cloud or dense depth. At its core is BoxerNet, a transformer-based network which lifts 2D bounding box (2DBB) proposals into 3D, followed by multi-view fusion and geometric filtering to produce globally consistent de-duplicated 3DBBs in metric world space. Boxer leverages the power of existing 2DBB detection algorithms (\textit{e.g.,} DETIC~\cite{zhou2022detecting}, OWLv2~\cite{minderer2023scaling}, SAM3~\cite{carion2026sam3}) to localize objects in 2D. This allows the main BoxerNet model to focus on lifting to 3D rather than detecting, ultimately reducing the demand for costly annotated 3DBB training data. Extending the CuTR~\cite{lazarow2024cubify} formulation, we incorporate an aleatoric uncertainty for robust regression, a median depth patch encoding to support sparse depth inputs, and large-scale training with over 1.2 million unique 3DBBs. BoxerNet outperforms state-of-the-art baselines in open-world 3DBB lifting, including CuTR in egocentric settings without dense depth (0.532 vs. 0.010 mAP) and on CA-1M with dense depth available (0.412 vs. 0.250 mAP). Project page with code available here: \url{https://facebookresearch.github.io/boxer}.

\end{abstract}

\section{Introduction}\label{sec:intro}

The transition from passive observation to active interaction in contextual AI and embodiment requires engaging the metric 3D world. 
% commonly referred to via human semantic concepts.
Estimating 3D bounding boxes (3DBBs) for arbitrary objects is central to this process, providing the semantic and geometric foundation for spatial understanding. 
An agent must reason about object identity, scale, orientation, and proximity to support goal-directed autonomous navigation and dexterous manipulation.
The same capability is critical for immersive augmented reality (AR) and digital twins, where AI must align semantic knowledge with physical space for seamless human-machine collaboration.

Despite being a cornerstone of contextual AI and embodiment, 3DBB estimation remains fraught with unresolved challenges. The primary obstacle is a profound data-annotation disparity. While 2D detection scales through intuitive image-space labeling~\cite{lin2014coco, kuznetsova2020open}, metric 3DBB estimation from a monocular view is inherently ill-posed. Collecting and annotating 3DBBs therefore requires specialized setups, typically involving LiDAR, depth sensors, or calibrated multi-view systems for 3D grounding. This creates a bottleneck, leaving even the largest 3D datasets orders of magnitude smaller than their 2D counterparts~\cite{brazil2023omni3d, lazarow2024cubify}. As a result, 3D models are often starved of the ``web-scale'' diversity that has enabled 2D vision-language models to learn the long tail of open-world objects.
Furthermore, the current 3DBB data landscape is fragmented across sensors and modalities. Available datasets form a patchwork: some provide dense depth maps from RGB-D cameras, while others provide sparse point clouds from simultaneous localization and mapping (SLAM)~\cite{aria2024tool} or structure-from-motion (SfM)~\cite{schonberger2016structure}. Data are also captured with diverse camera models, ranging from pinhole to fisheye and panoramic lenses.

The data and annotation landscape leaves the field under-explored and technologically rigid. Some methods remain trapped in a ``closed-set'' taxonomy, with no clear path to the open-world long tail~\cite{yang2025threedmood,zhang2025detany3d}, as they rely heavily on closed-set 3D supervision~\cite{brazil2023omni3d}. Other approaches rely heavily on 3D supervision for both detection and localization, placing high demands on training data and failing to leverage mature 2D detection as a bootstrap~\cite{straub2024efm3d, lazarow2024cubify}. Moreover, current 3D estimation paradigms struggle to handle this heterogeneity because camera models and sensor modalities are often baked into the architecture, requiring substantial re-engineering or retraining to transfer across settings. In summary, there is still no universal interface for 3D perception that bridges powerful 2D semantics and metric 3D geometry. Such a system should adapt its predictions to available geometric cues, whether from dense geometry or sparse points, while remaining agnostic to the underlying camera model.
% \fi
% --------------- I'm cutting line ---------------

% \lingni{CA-1M and CUTR are both from the same paper right? which corresponds to reference 15. reference 10 should be replaced with 15?}
% dd -- good catch, fixed now

% \lingni{The current writing only mention lack of data and difficulty in annotation. I'd suggest we also mention on why the few existing algorithms are not sufficient -- either because the design demand large 3D supervision or performance is not satisfying because only small-scale experiments are reported, or no open-world consideration, or no cross data generalization etc.}

% where is gap in existing 3DBB approaches -- limited data? combining data across different camera models, requiring dense depth. formulation relaxes requirements on dataset applicability. has not been shown yet.

% Then the next paragraph is spent on briefly motivating the design, provide the rational to factor out 2D detection and 3D lifting, and motivating multiview fusion.

Motivated by this gap, we bridge 2D and 3D detection by decomposing the problem into two stages: 2D localization followed by metric 3D lifting. By using existing open-world 2D detectors trained on internet-scale data, we inherit broad semantic coverage and strong localization in the image plane. We then focus model capacity on the geometric lifting step, learning to predict metric 3D bounding boxes from 2D proposals and depth cues. As illustrated in Fig.~\ref{fig:teaser}, this decomposition allows us to benefit from the scale of 2D supervision while addressing the geometric requirements of 3D detection. Additionally, we use flexible input conditioning to support training on pinhole and fisheye camera models and on datasets with and without sparse point clouds or dense depth maps. The contributions of this work are summarized as follows.

% \js{the previous paragraph and the bullet points after this one say pretty much the same thing} We present an approach for estimating any 3D object in a scene called Boxer. We focus on open-world detection, which aims to detect any object that can be described with natural language. Since the quantity of annotated 3DBB datasets is significantly less than that of 2DBB datasets from internet scale data, we propose to first use a 2DBB detector to localize the objects in the video in 2D, then use a learned estimator to find their 3D position relative to the known calibrated camera pose of the egocentric capture device.
% \js{and the following also communicates the same}

% We make the following contributions in this paper:
\begin{itemize}
  \item We propose \textbf{\boxer}, a complete algorithm for estimating open-world, global, de-duplicated 3D bounding boxes for objects using posed, calibrated video with optional sparse point clouds or dense depth.
    % \fan{also include the point that works well for (hopefully) any sensor suites that contains dense or sparse depth information?}

  \item We introduce \textbf{BoxerNet}, a transformer-based model that lifts 2D bounding box detections from off-the-shelf open-world detectors into metric 3D bounding boxes. It improves upon the CuTR~\cite{lazarow2024cubify} architecture by incorporating an \textbf{aleatoric uncertainty} head and a \textbf{median depth patch encoding} which enables \textbf{large-scale training} on over 1.22 million unique 3DBBs across four different device types.

  \item BoxerNet \textbf{outperforms state-of-the-art baselines} in the task of open-world 2D to 3D bounding box lifting across different camera types, including outperforming CuTR in egocentric settings without dense depth (0.532 vs. 0.010 mAP) and on CA-1M with dense depth available (0.412 vs. 0.250 mAP).

    % \fan{why only mention CuTR, probably should be "SOTA 3D detection algorithms"?}

  \item We release code and model to enable inference on a variety of input data.

\end{itemize}

\section{Related Work}\label{sect:related}

We review closed-set and open-set 3D object detection methods as well as, briefly, open-set 2D bounding box detection since Boxer builds on these.

\myparagraph{Closed-world 3D detection.} Many approaches for 3DBB detection exist that focus on a fixed set of classes (e.g. chair, table, lamp) such as Cube-RCNN~\cite{brazil2023omni3d} which uses a Faster-R-CNN-style 2D detection approach on a large set of annotated data called Omni3D to detect objects from monocular images. Other works such as H3DNet~\cite{zhang2020h3dnet}, FCAF~\cite{fcaf3d2022}, TR3D~\cite{tr3d2023}, VoteNet~\cite{qi2019votenet} and 3DETR~\cite{misra2021detr} rely primarily on sparse point clouds input from a depth scan to detect a fixed set of classes. EVL~\cite{straub2024efm3d} operates on a multi-camera stereo rig such as Project Aria~\cite{aria2024tool,aria_gen2_pilot} and follows a similar detection strategy as ImVoxelNet~\cite{rukhovich2022imvoxelnet} that lifts 2D features into a 3D voxel grid. More recent Transformer-based multi-view methods like DETR3D~\cite{wang2022detr3d}, PARQ~\cite{xie2023parq} or PETR~\cite{liu2022petr}, directly query 3D objects from calibrated image sets.  SceneScript~\cite{avetisyan2024scenescript} formulates 3DBB estimation in an autoregressive way using a custom object language. SpatialLM~\cite{SpatialLM} is another example of a 3D-LLM capable of detecting objects in 3D, but detection is still limited to a closed set of 59 common categories and trained only on synthetic data. While the aforementioned works focus on indoor scenes, 3D detection for autonomous driving typically adopts Bird’s-Eye-View (BEV) approaches~\cite{philion2020lift,li2022bevformer,li2023bevdepth} to localize fixed-set classes like cars and pedestrians.

\myparagraph{Open-world 2D detection.} Open-world 2D object detection is a well studied, extensive field. Here we focus only on the key most relevant approaches that take an input text prompt such as ``red car'' and localize it in the image with a 2D bounding box and optionally a segmentation mask. DETIC~\cite{zhou2022detecting} is an example that is capable of detecting 21k+ categories that used weak annotations from image level annotations. Powerful models such as OWLv2~\cite{minderer2023scaling} can detect tens of thousands of classes and relies on pseudo annotation to self-label over 1 billion weakly annotated web image-text pairs. SAM3~\cite{carion2026sam3} is another approach which uses a large and diverse corpus of 4 million object-text concepts and includes segmentation masks. Vision–language models (VLMs) like Qwen3-VL~\cite{bai2025qwen3} do not mention an exact number of 2D bounding box annotations but are pretrained on web-scale image–text corpora involving hundreds of billions of tokens, which supports robust multimodal grounding and open-vocabulary reasoning.

%  Most recent VLMs are able to output 2D bounding box grounding information and are also trained on over one billion image-text pairs.

% \js{here it seems like the 2D BB datasets are not that big either? in the intro you talk about 2 orders of magnitude difference}

\myparagraph{Open-world 3D detection.} Open-world 3DBB detection is a less solved field. A promising shift in open-set 3D detection treats localization as a generative ``next-token prediction'' task~\cite{cho2024language}, leveraging the broad semantic knowledge and reasoning capabilities of VLMs. Building on this, LocateAnything3D~\cite{man2025locateanything3d} introduced a Chain-of-Sight (CoS) decoding strategy that mirrors human reasoning by emitting 2D grounding tokens as a visual anchor before predicting 3D coordinates. While large-scale foundation models like Qwen3-VL~\cite{bai2025qwen3} and N3D-VLM~\cite{wang2025n3d} have recently incorporated 3D grounding into their unified multimodal output via JSON-like tokens, their performance is often hampered by 3D hallucinations and a lack of metric precision compared to dedicated geometric systems.

Similar to Boxer, 3D-MOOD~\cite{yang2025threedmood} and DetAny3D~\cite{zhang2025detany3d} also lift open world 2D detections to 3D conditioned on monocular images. One limitation of these approaches is that they cannot condition on commonly available known depth from a depth sensor or sparse point clouds because they compute dense depth internally. OVMono3D~\cite{yao2024open} proposes a framework for open-vocabulary monocular 3D detection that decouples 2D semantic recognition from 3D spatial estimation, leveraging pretrained vision foundation models to lift 2D bounding boxes into metric 3D space for novel categories. These approaches focus on monocular detection, whereas Boxer supports an optional metric scale world coming from a calibrated stereo rig or RGB-D sensor.
The closest to Boxer is CuTR~\cite{lazarow2024cubify}, which trains a DETR-style 3DBB detector using open-world 3DBB annotated data from ARKit Scenes~\cite{baruch2021arkitscenes}. In Boxer we follow a similar architecture to CuTR, but we replace the pixel level dense depth ViT with a more flexible depth patches encoder to enable scaling to different depth input types. Since CuTR is built on the DETR framework, a confidence score is available via the foreground/background classifier score. However, this couples both the 2D and 3D detection scores. In our work we factorize this into a separate 3D uncertainty.
SAM3D~\cite{chen2025sam} goes a step beyond 3DBB detection (which they described as layout estimation) and estimates a full 3D shape plus texture of each object. Combined with an open-world detection and segmentation model such as SAM3~\cite{carion2026sam3}, it can serve a similar function to Boxer, while lacking the ability to process depth input and fuse estimates from multiview or videos.

To bridge the gap between 2D reasoning and 3D geometric consistency, recent works such as ConceptGraphs~\cite{gu2024conceptgraphs} and EgoLifter~\cite{gu2024egolifter} lift open-world 2D segmentation masks into 3D, constructing object-centric scene representations by associating class-agnostic segments across multi-view sequences. In contrast, Boxer functions as a flexible detector designed for efficient 3DBB estimation, with the post-processing temporal fusion step to achieve 3D consistency.

\section{Boxer}
\label{sect:boxer}

\begin{figure}[t]
  \centering
  \includegraphics[width=0.99\textwidth]{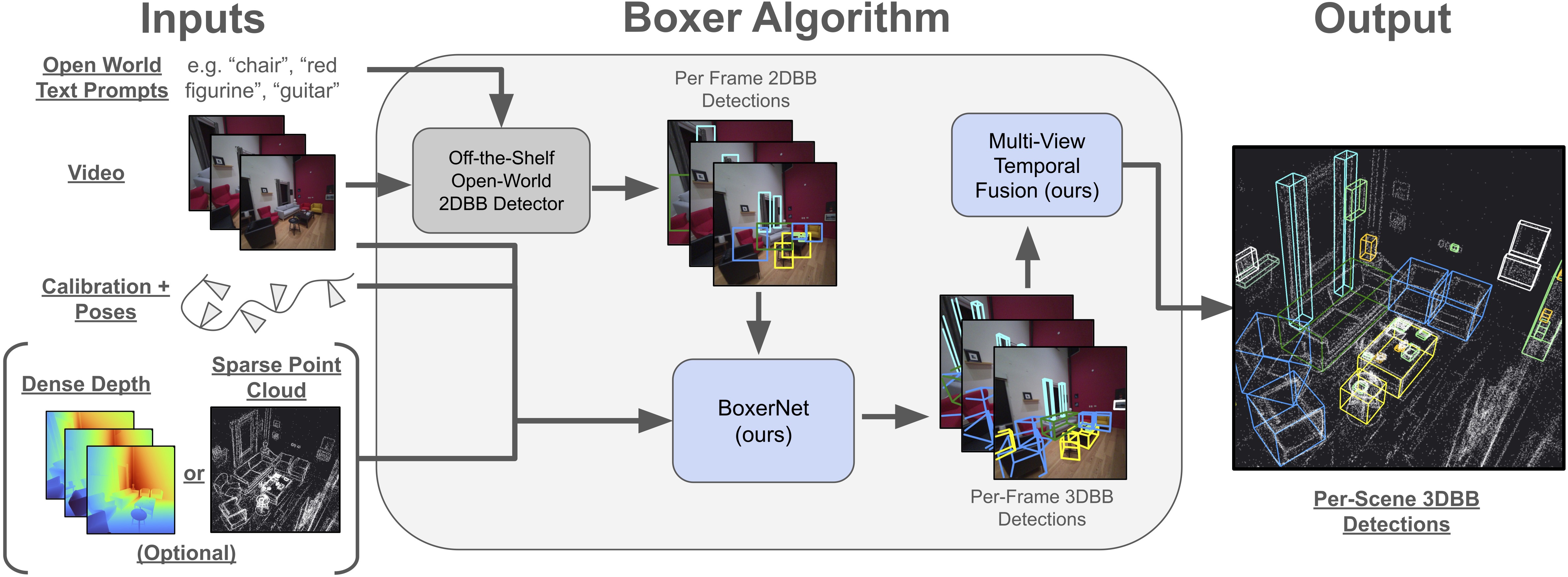}
  \caption{\textbf{Boxer algorithm overview.}
  Boxer operates on a set of posed and calibrated images with optional dense depth or sparse point cloud to produce metric, static, 3D bounding boxes for open-set objects.}
  \label{fig:system}
\end{figure}

Boxer is an algorithm that takes as input a set of posed, calibrated images and produces a final set of global, metric 3DBBs. As shown in Fig.~\ref{fig:system}, BoxerNet (Sec.~\ref{sec:boxernet}) is the core module and produces per-frame 3DBB detections from a single image and open-world 2DBBs from an off-the-shelf detector (Sec.~\ref{sec:2d_det}). BoxerNet can also take optional depth input to improve performance.
Per-frame 3DBB detections are merged and filtered using a hand-engineered fusion algorithm (Sec.~\ref{sect:fusion}).
% \lingni{consider have a short section to define all the math symbols}

\subsection{2D Detection}
\label{sec:2d_det}
The Boxer pipeline first uses an off-the-shelf open-world 2D bounding-box (2DBB) detector to localize text queries in the image. We denote by $\mathcal{T} = \{ t_j \}_{j=1}^M$ a set of natural-language text prompts used to query an open-world 2D bounding-box detector. Given an image $I \in \mathbb{R}^{H \times W \times 3}$ and a prompt $t_i$, the detector produces a set of 2D bounding boxes $\mathcal{B}^{2D} = \{ b_i \}_{i=1}^{N}$ corresponding to image regions semantically aligned with the query, since we can have more than one detection for a given text query. Note we trivially handle grayscale images by repeating the single channel three times. Each box $b_i$ is parameterized by its corner coordinates and an associated 2D confidence score $s_i^{2D} \in [0,1]$.

% \js{x and y here collide with the 3D x, y, z in next subsection. Do people need to know 2D BB parameterization?}
% dd -- good point, I dropped it to avoid confusion

% \js{only introduce notation that is necessary - are you using $H$ and $W$ anywhere?}
% dd - yes we need H,W later on to explain H' and W'

% \js{$i$ and $j$ not introduced clearly - are they needed? I think we may not need $j$ - every 2D BB stems from a class/text prompt $t_i$}
% dd - we need j and i (and M vs N) since a single text prompt can have multiple detections, but we can drop j past this point for simplicity

\subsection{BoxerNet: Lifting 2D to 3D}
\label{sec:boxernet}

\begin{figure}[!t]
  \centering
  \includegraphics[width=0.98\textwidth]{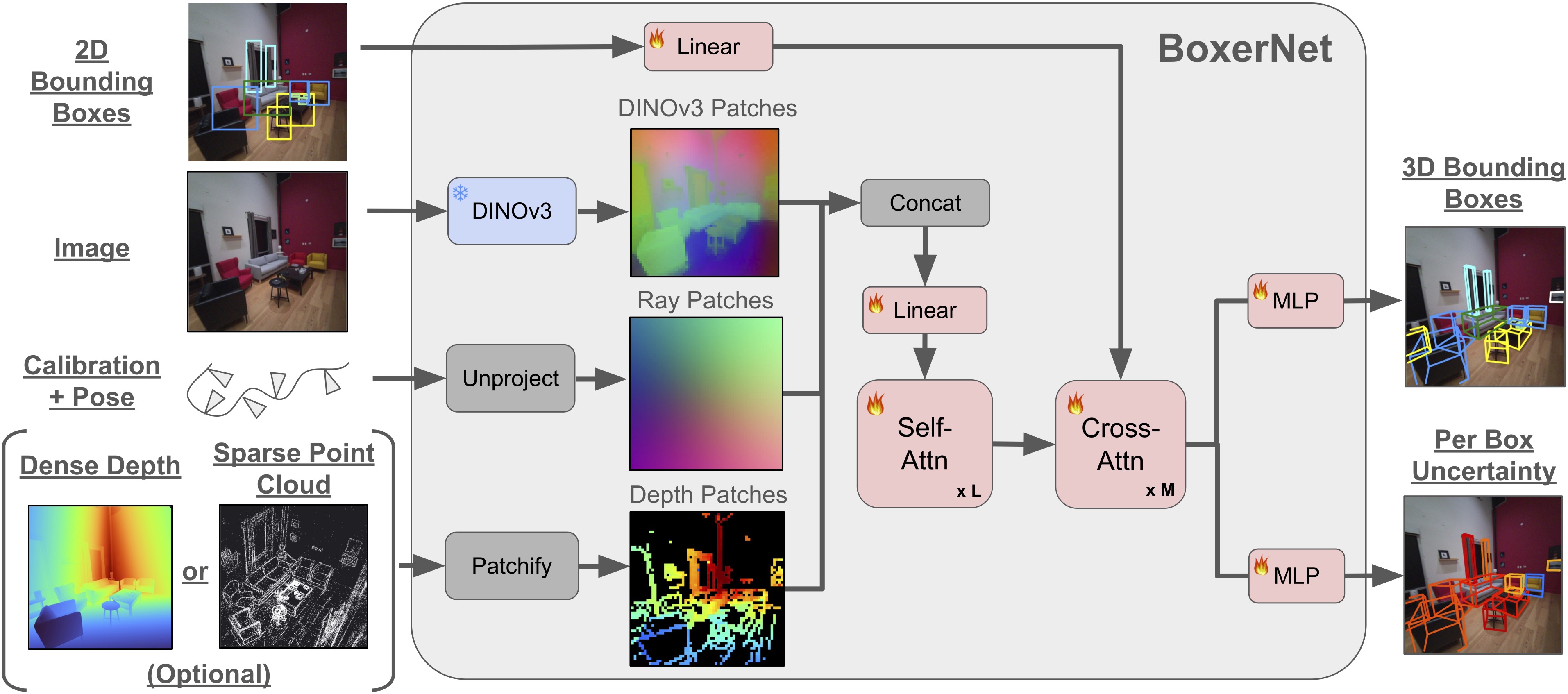}
  \caption{\textbf{BoxerNet lifting module.}
  BoxerNet conditions on the image, camera calibration and poses and an optional depth input to lift 2D bounding boxes into metric 7-DoF 3D bounding box predictions.}
  \label{fig:boxernet}
\end{figure}

\myparagraph{Overview.} We show the BoxerNet network design in \cref{fig:boxernet}. Each 2D detection $b_i^{2D} \in \mathcal{B}^{2D}$ is mapped to a 3D detection $b_i^{3D} \in \mathcal{B}^{3D}$ using a learned model
$f_{\text{lift}}$ that predicts a 3D object hypothesis conditioned on the image and the 2D box.
Specifically, the model outputs a 7-DoF 3DBB
\begin{equation}
  b_i^{3D} = (x_i, y_i, z_i, w_i, h_i, d_i, \theta_i) \;\;,
\end{equation}
% \lingni{should be $b_i^{(j)}$ right? I find the notation a bit complicated. Maybe first define $\boldsymbol{b} = (x, y, z, w, h, d, \theta)$ to denote 3DBB paramterization, then $\boldsymbol{b}_{ij}$ to refer the $i$-th 3DBB in $j$-th image.}
% dd -- check it again, is it better now? I dropped all the j notation at this point
% \lingni{why is it in the world?}
% dd - good catch, updated to gravity aligned camera coordinate frame
% \js{x, y already used for 2D BBs}
% dd - removed x,y, from 2DBBs
where $(x_i, y_i, z_i) \in \mathbb{R}^3$ denotes the object center in the gravity-aligned camera coordinate frame $\mathcal{F}_g$,
$(w_i, h_i, d_i) \in \mathbb{R}_{+}^3$ its physical extent, and $\theta_i \in [-\pi, \pi)$ a single rotation
about the gravity axis.
Each lifted prediction is additionally associated with a confidence score
$s_i^{3D} \in [0,1]$, indicating the model's confidence in the 3D hypothesis. We choose a 7-DoF representation because the datasets commonly provide a gravity estimate via an IMU, but it is straightforward to extend this to a 9-DoF 3DBB representation (\ie estimate a quaternion instead of a single scalar).

\myparagraph{Lifting encoder.} The lifting model $f_{\text{lift}}$ is implemented as a self-attention encoder that jointly reasons over appearance, scene depth, and camera calibration. Given an input image $I$, we extract dense visual features
$F^{\text{img}} \in \mathbb{R}^{H' \times W' \times D}$
using a pretrained DINOv3 \cite{simeoni2025dinov3} backbone. To encode metric scale into the model, we optionally project sparse point clouds or dense depth points from a depth sensor into the image plane using known camera intrinsics and compute the median depth within each image patch, yielding a per-patch depth feature $F^{\text{depth}} \in \mathbb{R}^{H' \times W' \times 1}$. If no point projects into a patch, we set that patch to -1.
Camera calibration and pose information are encoded as ray features
$F^{\text{ray}} \in \mathbb{R}^{H' \times W' \times 3}$ following \cite{wang2022input},
where each ray encoding represents the normalized 3D ray associated with an image patch in the gravity-aligned coordinate frame $\mathcal{F}_g$, computed by unprojecting the patch center. Camera translation is omitted because it is always set to the origin $(0,0,0)$.
Sparse point clouds with per-image visibility are readily available in SLAM or SfM systems \cite{aria2024tool,campos2021orbslam3,schonberger2016structure}. This differs from CuTR, which requires a specialized ViT that processes dense depth only.

If 6-DoF pose is not available, the 2-DoF gravity direction (obtained by an IMU or off-the-shelf estimator \cite{veicht2024geocalib}) could be used instead of the full pose for single-image box lifting, though all datasets we tested with provide the full pose.
The three feature modalities are concatenated per patch $F^{\text{enc}} \in \mathbb{R}^{H' \times W' \times D + R + 1}$ and fed as tokens into a shared self-attention encoder which allows the semantic features to mix with the ray and optional depth values. Including the rays and depth as separate channels enables conditioning on intrinsics with or without depth.

\myparagraph{2D-to-3D box decoder.}
Following self-attention encoding, the lifted 3D predictions are obtained by conditioning on the
2D detections via cross-attention.
Each 2D box $b_i^{(j)}$ is first lifted to the same feature dimension as the patch tokens using a linear
layer (4 to hidden dimension), and then cross-attended over all encoder tokens (\ie all patches), producing
a box-specific latent representation. Each box attends independently to the patch tokens, with no attention
between box tokens, making the formulation permutation invariant. This representation is passed to two prediction heads.  Each head is a two-layer MLP with ReLU and 128 hidden dimensions.

\myparagraph{Outputs.}
The first head regresses the 7-DoF 3D bounding box parameters
$\hat{b}_i^{3D} = (\hat{x}_i, \hat{y}_i, \hat{z}_i, \hat{w}_i, \hat{h}_i, \hat{d}_i, \hat{\theta}_i)$.
The second head predicts an aleatoric uncertainty value
$\hat{\sigma}_i$~\cite{kendall2017uncertainty}, which captures observation noise and ambiguity in the lifting process. The final confidence score $s_i$ for each box is the mean of the 2D and 3D confidences:
\begin{equation}
  s_i = (s_i^{2D} + s_i^{3D}) / 2 \;\;,
\end{equation}
which is used during inference to rank the final detections for filtering and generating the PR curves needed in the mAP computation.
% \js{I remember you did some studies to get to this way of fusing - worth mentioning if have space. especially since other works combine them differently i.e. CubeRCNN}

% \js{using a XY activation}
% dd -- I don't know what you mean by that

\myparagraph{Training objective.}
The lifting model is trained using a loss function that combines geometric alignment with aleatoric uncertainty~\cite{kendall2017uncertainty} modeling. Specifically, we supervise the predicted 3D bounding box using a Chamfer loss
$\mathcal{L}_{\text{chamfer}}$ between the predicted and ground-truth box corners. We also add the 3D uncertainty term as in~\cite{brazil2023omni3d,lu2021gupnet}. The final training objective is given by
\begin{equation}
  \mathcal{L}
  =
  \mathcal{L}_{\text{chamfer}} \cdot \exp(-\hat{\sigma})
  + \hat{\sigma} \;\;,
\end{equation}
where $\hat{\sigma}$ denotes the predicted aleatoric uncertainty from above.

\subsection{Multi-View Temporal Fusion}
\label{sect:fusion}

The final step estimates a set of per-scene de-duplicated, global, static 3DBBs from per-frame 3DBB estimates across time and views. Per-frame 3D detections are aggregated into a consistent set of scene-level object hypotheses via a geometric and semantic fusion pipeline.
Given the set of lifted 3D bounding boxes across time, merging is first restricted to geometrically compatible detections using a 3D IoU threshold, preventing fusion of spatially distant objects.
Semantic consistency is enforced by allowing merges only between detections with similar category labels computed on the prompt text via a text embedding \cite{sbert}.
% \js{how is semantic similarity measured, do you hard threshold as well to get graph?}
A graph is then constructed whose nodes correspond to detections and whose edges connect geometrically and semantically compatible pairs. Connected components define object-level clusters spanning multiple frames.
Within each cluster, confidence-weighted fusion of box parameters is performed. To account for the $90^\circ$ rotational symmetry of gravity-aligned boxes, orientation ambiguity is resolved before averaging position, dimensions, and yaw (using a circular mean). Finally, 3D non-maximum suppression is applied to the fused hypotheses to remove residual duplicates, yielding a compact set of scene-level 3D object detections.
Additional implementation details are provided in the supplementary material.

\section{Training Dataset}
\label{sect:dataset}

\myparagraph{Overall statistics.} We train Boxer on both internal and public data. The internal non-public dataset consists of Project Aria (Gen1~\cite{aria2024tool} and Gen2~\cite{aria_gen2_device_2025}) and Quest3 data. For public data, we use NymeriaPlus~\cite{nymeriaplus}, which has additional annotation on top of~\cite{nymeria} and CA-1M~\cite{lazarow2024cubify}. We also source data from Omni3D by including SUN-RGBD and ScanNet~\cite{dai2017scannet} (using 3DBBs from Scan2CAD~\cite{avocado2019scan2cad}). The included datasets are shown in \cref{tab:dataset_comparison}. We focus on indoor datasets because they are an important use case for everyday human and robotic applications. We exclude other Omni3D indoor datasets: ARKit~\cite{baruch2021arkitscenes} due to overlap with CA-1M, Objectron~\cite{objectron2021} due to its single-object focus, and Hypersim~\cite{hypersim} because it is synthetic. Summed across these sources, this yields roughly 1.22M unique 3DBBs and 42.1 million image views.

% \lingni{add categories?}
\begin{table}[bt]
  \centering
  \scriptsize % Slightly smaller font to ensure it fits the column width
  \renewcommand{\arraystretch}{1.05}
  \setlength{\tabcolsep}{8pt} % Add a bit more horizontal space between columns
  \caption{\textbf{Comparison of datasets used for developing Boxer.} We use both internal and public data sources. \vspace{-1ex}}
  \label{tab:dataset_comparison}
  \begin{tabular}{@{}lllcc@{}}
    \toprule
    \textbf{Dataset Name} & \textbf{Device} & \textbf{Availability} & \textbf{Unique 3DBB} & \textbf{Images} \\ \midrule
    -                   & Project Aria & Internal & 65k & 1.1M \\
    -                   & Quest        & Internal & 400k & 6.4M \\
    NymeriaPlus         & Project Aria & Public   & 240k & 8.6M \\
    CA-1M               & iPad         & Public   & 440k & 13M \\
    ScanNet             & iPad         & Public   & 14k & 2.5M\\
    SUN-RGBD            & Kinect       & Public   & 65k & 10.5k\\
    \bottomrule
  \end{tabular}
\end{table}

% \js{this kind of data would be most easily presented in a table. then just mention the overal stats from the last sentences}

Note that we focus on \emph{unique} 3D bounding boxes when quantifying the scale of such datasets. This is because with posed video datasets, it is trivial to record a long static video, annotate a single 3DBB, and annotate a thousand 3DBB \emph{observations}. Thus we separate the statistics into unique 3DBBs and unique image views.

% However, even with different viewpoints, this does not provide sufficient diversity to teach ML models\js{strong statement - reviewers might question}.
% \js{I called them observations above - okay?}.

% dd - note that observations is different than images, one image can have many observations. I don't have the observation stat handy.

% \begin{figure}[!ht]
% \centering
% \includegraphics[width=0.98\textwidth]{figures/misc/datasetv2.jpg}
% \caption{\textbf{Example 3D GT Annotations.}
% Some examples of the 3D ground truth annotations across different cameras are shown.}
% \label{fig:dataset}
% \end{figure}

\myparagraph{Per-view visibility.} Since per-view 2D annotation is costly and the training scenes are static, some datasets only have static 3D bounding boxes with human annotation.
Naively projecting all 3D bounding boxes into each image does not account for occlusion (\ie boxes will be visible through walls).
To compute the per-frame visibility of each 3D bounding box with respect to each posed image, we use the observability of the available depth, making sure at least two visible depth points lie inside the 3DBB.
In addition, we require sufficient visibility of the 3D bounding box geometry by enforcing that 80\% of sampled points along the box edges are visible in the valid region of the image.

% More formally, let $\mathcal{P}_f$ denote the set of semi-dense 3D points observed in frame $f$,
% and let $\mathcal{E}(B) = \{e_i\}_{i=1}^N$ denote a set of points sampled uniformly along the edges of a 3D bounding box $B$.
% A 3D bounding box $B$ is considered visible in frame $f$ if
% \[
% \sum_{p \in \mathcal{P}_f} \mathbb{I}(p \in B) \ge 2
% \quad \text{and} \quad
% \frac{1}{|\mathcal{E}(B)|}
% \sum_{e \in \mathcal{E}(B)} \mathbb{I}(e \in \mathcal{P}_f) \ge 0.2 .
% \]
% \js{can you just say it in english?}

% \textbf{Model-in-the-Loop Annotation}.
% We integrated Boxer into our internal annotation tool, to speed up annotation. Similar to SAM \cite{kirillov2023segment}, annotators initialize a new box using a 2D bounding box prompt which is input to Boxer. On a small test sequence of 30 objects, the model in the loop annotation reduces annotation time of an expert annotator from 38 to 20 sec per box, roughly a \textbf{1.8x speed up}.

\myparagraph{Data augmentation.} We apply four types of data augmentation to the encoder inputs to improve robustness and generalization: photometric, camera, depth, and 2D box augmentation. See the supplemental material for details.

\myparagraph{Training and inference details.}
The model is trained on the dataset described in \cref{sect:dataset}. It is trained for roughly two weeks on 16 H100 GPUs using the AdamW optimizer with a cosine-decaying learning rate that starts at 1e-4 and decays to 1e-5. BoxerNet lifting, given 2DBB detection, takes roughly 20 ms for a forward pass on $960 \times 960$ images using bfloat16 on an NVIDIA RTX 4090 and has about 25 million trainable parameters.

\section{Experiments}
\label{sect:experiments}

% \js{why state that? seems obvious?}
% We focus the evaluation on BoxerNet, as that is the key contribution of this work.

Our experiments are split into three parts: (1) per-frame single-image evaluation with optional depth (dense depth or sparse point clouds), which computes metrics on 3DBBs predicted independently for each frame; (2) per-scene evaluation, which runs the fusion described in \cref{sect:fusion} on each method and computes metrics on the final set of estimated 3DBBs; and (3) sensitivity studies to analyze the key factors in Boxer's design.

\subsection{Testing Datasets}

\paragraph{NymeriaPlus~\cite{nymeriaplus}.} The NymeriaPlus dataset provides a test set that is exhaustively annotated with open-world 3D bounding boxes. The data comes from real-world homes captured by egocentric glasses. This dataset does not have external ground-truth sensors (e.g., no dense depth). Thus, we can only compare methods that operate on image-only input or image + sparse point clouds from the SLAM system.

\paragraph{Aria Digital Twin~\cite{aria_digital_twin}.} We use the Aria Digital Twin (ADT) dataset, which contains ground-truth depth, to compare models that can use depth as input (e.g., CuTR (Image-D)). This dataset contains one indoor apartment scene collected from egocentric data. We select a single sequence from this dataset. The sequence includes a small amount of dynamics (\textless{}5\% of observed 3DBBs). We exclude such dynamic objects from evaluation. For GT2D, we prompt only with 2DBBs from static objects.

\paragraph{CA-1M~\cite{lazarow2024cubify}.} We use the first ten sequences from the validation set to measure performance on CA-1M. We use the provided rendered/cropped 3DBBs for testing, as well as provided 2DBBs for GT2D prompt inputs. We exclude very large structural objects such as floors and walls, or any object with a dimension > 3 m, as these are typically not observable in a single image. We additionally expand the dimensions of thin objects (e.g., lights) to at least 5 cm for both predictions and ground truth.

\paragraph{Omni3D~\cite{brazil2023omni3d}.} We also report results for our model on the closed-world Omni3D dataset, focusing on the SUN-RGBD split. We use the provided 2D boxes for GT2D prompts, as well as provided dense depth for depth-based models.

\subsection{Benchmarking Protocols}
\myparagraph{Open-world 2DBB detectors.}
We primarily experiment with two open-world 2DBB detectors: DETIC \cite{zhou2022detecting} and OWLv2 \cite{minderer2023scaling}. We also evaluate SAM3 \cite{carion2026sam3}, but its compute cost is high with 1000+ prompts, taking 45s per image on NVIDIA RTX 4090, versus 35ms for DETIC and 120ms for OWLv2, making it less suitable for long sequences with thousands of images. We also use ground-truth 2D bounding boxes (GT2D) in some experiments to focus evaluation on the 3D lifting capability of each method.

\myparagraph{Text prompts.}
For open-world datasets (NymeriaPlus, ADT, and CA-1M), we use a generic open-world prompt consisting of LVIS (1202 categories) plus 17 additional common indoor-focused categories. See the supplemental material for the exact list. In Omni3D, which is a closed-set 3DBB dataset, we prompt open-world models with the limited taxonomy of 50 classes, as done in~\cite{brazil2023omni3d}.

% \subsection{SAM3D Comparison}
% \label{sec:sam3d}

% \begin{table}[!ht]
%   \centering
%   \tighttable
%   \scriptsize
%   \setlength{\tabcolsep}{8pt}
%   \renewcommand{\arraystretch}{1.05}
%   \caption{\textbf{SAM3D comparison on ADT (RGB-only).} We report class-agnostic per-frame mAP for three RGB methods on ADT.\vspace{-1ex}}
%   \begin{tabular}{l c}
%     \toprule
%     Method & ADT \\
%     \midrule
%     OWLv2+BoxerNet & \TODO{} \\
%     SAM3+BoxerNet  & \TODO{} \\
%     SAM3+SAM3D     & \TODO{} \\
%     \bottomrule
%   \end{tabular}
%   \label{tab:sam3d_adt}
% \end{table}

\begin{table}[!ht]
  \centering
  \tighttable
  \scriptsize
  \setlength{\tabcolsep}{8pt}
  \renewcommand{\arraystretch}{1.05}
  \caption{\textbf{Per-frame 3D detection results.} All numbers report class-agnostic mAP. Methods are grouped by Image-only and Image+Depth. For NymeriaPlus, since depth is not available, sparse point cloud from SLAM is used. \textbf{Bold} indicates the best method excluding GT2D and \underline{underline} the best lifting method for a given 2DBB input.\vspace{-1ex}}
  \begin{tabular}{l l | c c c c}
    \toprule
    & & \multicolumn{4}{c}{Test Set (mAP per-frame)} \\
    \cmidrule(lr){3-6}
    Method & Modalities & NymeriaPlus & ADT & CA-1M & $\mathrm{Omni3D}_{\mathrm{SUN}}$ \\
    \midrule
    %\multicolumn{5}{l}{\textit{Image Only}} \\
    %\midrule
    DETIC+BoxerNet  & RGB   & 0.110      & 0.026      & 0.053      & 0.098      \\
    \grayrow
    3D-MOOD       & RGB    & 0.014         & 0.003             & 0.046         & 0.271 \\
    \grayrow
    3D-MOOD+BoxerNet  & RGB   & \best{\underline{0.092}}  & \underline{0.013}      & \best{\underline{0.099}}  & \best{\underline{0.340}}\\
    CubeRCNN      & RGB   & 0.030      & 0.015      & 0.042      & 0.264  \\
    CubeRCNN+BoxerNet & RGB  & \best{\underline{0.092}} & \underline{0.030}      & \underline{0.075}      & \underline{0.286}  \\
    \grayrow
    % SAM3+SAM3D    & RGB   & \TODO{}      & \TODO{}      & \TODO{}      & \TODO{}     \\
    % SAM3+Boxer    & RGB   & \TODO{}      & \TODO{}      & \TODO{}      & \TODO{}      \\
    % \hdashline
    OWLv2+CuTR    & RGB   & 0.005  & 0.001  & 0.064  & 0.078      \\
    \grayrow
    OWLv2+BoxerNet   & RGB   & \underline{0.061}  & \best{\underline{0.041}}  & \underline{0.081}  & \underline{0.144} \\
    GT2D+CuTR    & RGB    & 0.010         & 0.004         & 0.119  & 0.160 \\
    GT2D+BoxerNet   & RGB    & \underline{0.296}  & \underline{0.077}  & \underline{0.126}  & \underline{0.293} \\
    \midrule
    %\multicolumn{5}{l}{\textit{Image + Depth}} \\
    %\midrule
    \grayrow
    DETIC+BoxerNet  & RGB+D   & 0.193      & 0.120      & 0.153  & 0.200      \\
    3D-MOOD+BoxerNet  & RGB+D  & 0.166      & 0.040      & 0.174      & 0.449 \\
    \grayrow
    CubeRCNN+BoxerNet & RGB+D  & 0.120  & 0.039      & 0.104      & \best{0.450} \\
    EVL          & RGB+D   & 0.100      & -         & -   & -   \\
    % EVL+Boxer    & RGB+D   & \TODO{}      & \TODO{}      & -   & -   \\
    % \hdashline
    % SAM3+SAM3D    & RGB+D  & -   & \TODO{}      & \TODO{}      & \TODO{}     \\
    % SAM3+Boxer   & RGB+D   &\TODO{}     & \TODO{}      & \TODO{}      & \TODO{}     \\
    \grayrow
    OWLv2+CuTR    & RGB+D  & -   & 0.052  & 0.178  & 0.160 \\
    \grayrow
    OWLv2+BoxerNet & RGB+D    & \best{\underline{0.297}}  & \best{\underline{0.124}}  & \best{\underline{0.204}}  & \underline{0.290}  \\
    GT2D+CuTR    & RGB+D   & -          & 0.102         & 0.250         & 0.219     \\
    GT2D+BoxerNet  & RGB+D    & \underline{0.532}  & \underline{0.317}  & \underline{0.412}  & \underline{0.585} \\
    \bottomrule
  \end{tabular}
  \label{tab:results_frame}
\end{table}

\begin{table}[!ht]
  \centering
  \tighttable
  \scriptsize
  \setlength{\tabcolsep}{8pt}
  \renewcommand{\arraystretch}{1.05}
  \caption{\textbf{Per-scene (fused per-frame) 3D detection results.} All numbers report class-agnostic mAP. Methods are grouped by Image-only and Image+Depth inputs (NymeriaPlus uses sparse point cloud from SLAM). We focus on video datasets and the CuTR baseline as it is most similar to BoxerNet and capable of running in real-time. \underline{Underline} indicates the best lifting method for a given 2DBB input.\vspace{-1ex}}
  \begin{tabular}{l l | c c c}
    \toprule
    & & \multicolumn{3}{c}{Test Set (mAP per-scene)} \\
    \cmidrule(lr){3-5}
    Method & Modalities & NymeriaPlus & ADT & CA-1M \\
    \midrule
    OWLv2+CuTR  & RGB & 0.013        & 0.004        & 0.064        \\
    OWLv2+BoxerNet & RGB & \underline{0.145} & \underline{0.077} & \underline{0.081} \\
    \grayrow
    GT2D+CuTR   & RGB & 0.013        & 0.005            & 0.087            \\
    \grayrow
    GT2D+BoxerNet  & RGB & \underline{0.219} & \underline{0.107}            & \underline{0.089}      \\
    \midrule
    OWLv2+CuTR & RGB+D  & -         & 0.198        & 0.178        \\
    OWLv2+BoxerNet & RGB+D & \underline{0.278} & \underline{0.215} & \underline{0.204} \\
    \grayrow
    GT2D+CuTR   & RGB+D & -         & 0.244            & 0.305            \\
    \grayrow
    GT2D+BoxerNet & RGB+D  & \underline{0.400} & \underline{0.307}            & \underline{0.434}      \\
    \bottomrule
  \end{tabular}
  \label{tab:results_scene}
\end{table}

\myparagraph{Metrics and protocol.} We report the performance of 3D detectors at the scene level after per-frame detections are fused into a global static 3D coordinate frame. We compute precision-recall curves for 3D IoU at thresholds [0.05, 0.1, 0.15, ..., 0.5] and average them to compute mAP, as done in prior work~\cite{brazil2023omni3d}. Unlike some prior setups, we do not limit detections to 100; instead, we run each detector at a relatively low threshold and report all resulting detections. We report class-agnostic results, meaning we discard semantic labels from each method and treat all boxes as a single ``Anything'' class.

\myparagraph{Explanation of combinations in \cref{tab:results_frame}.} To make comparisons as fair as possible, we focus evaluation on each model's ability to take a 2D box and lift it to 3D. BoxerNet does not provide 2D detections itself, so it requires 2DBB inputs. For methods that already use a 2DBB detector, this is straightforward (e.g., OWLv2+BoxerNet). For methods that are 3DBB detectors, such as 3D-MOOD~\cite{yang2025threedmood}, we first run their 3DBB detector, then project each 3DBB back into the 2D image using known camera pose and calibration to generate a 2DBB, and then prompt BoxerNet with it. We follow the same procedure for CubeRCNN. This reduces compounding variables such as taxonomy and training-set differences.

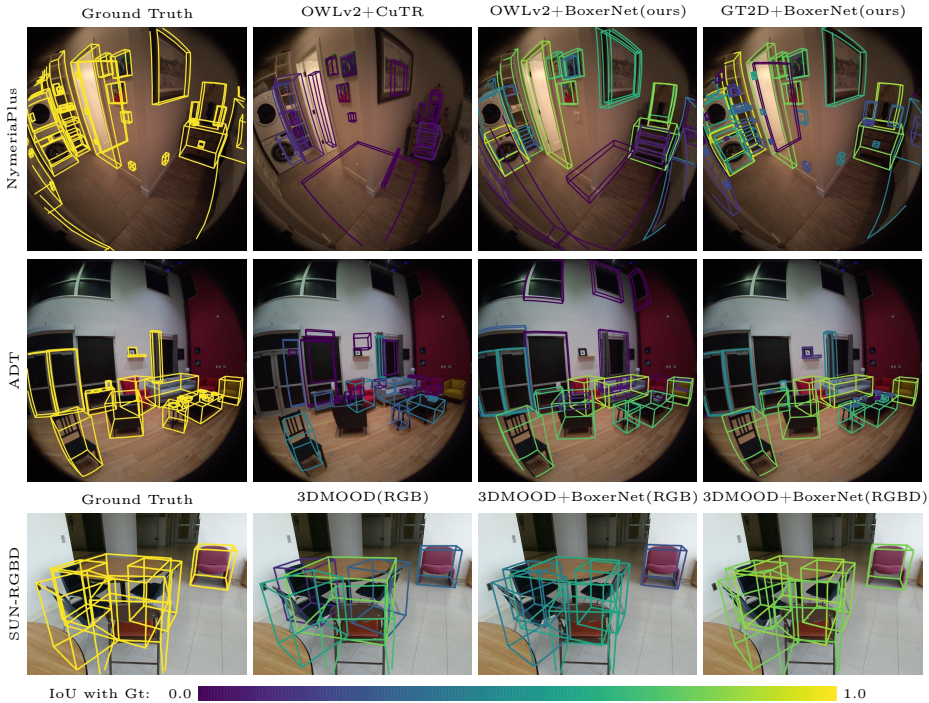
\begin{figure}[!tb]
  \centering
  \def\iw{9em}
  \def\dy{-1pt}
  \begin{tikzpicture}[inner sep=1pt]
    \node(p01)                            {\includegraphics[width=\iw]{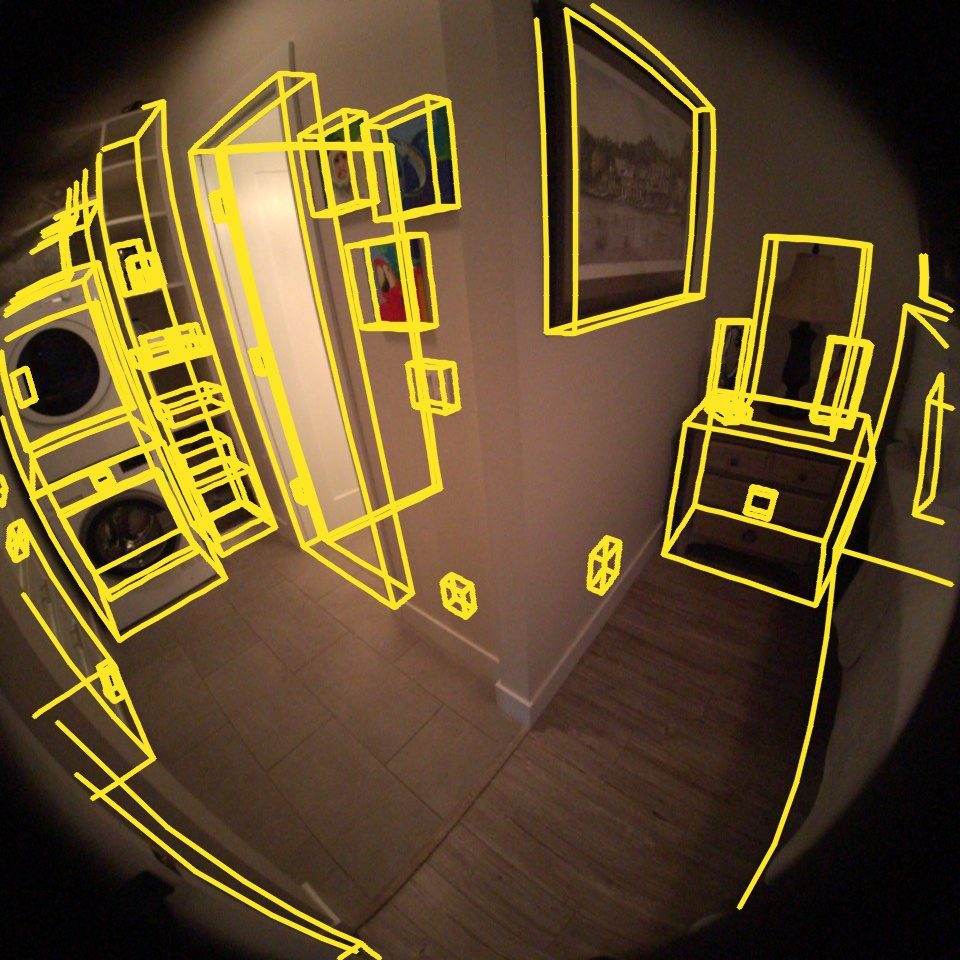}};
    \node(p02) at (p01.east) [anchor=west]{\includegraphics[width=\iw]{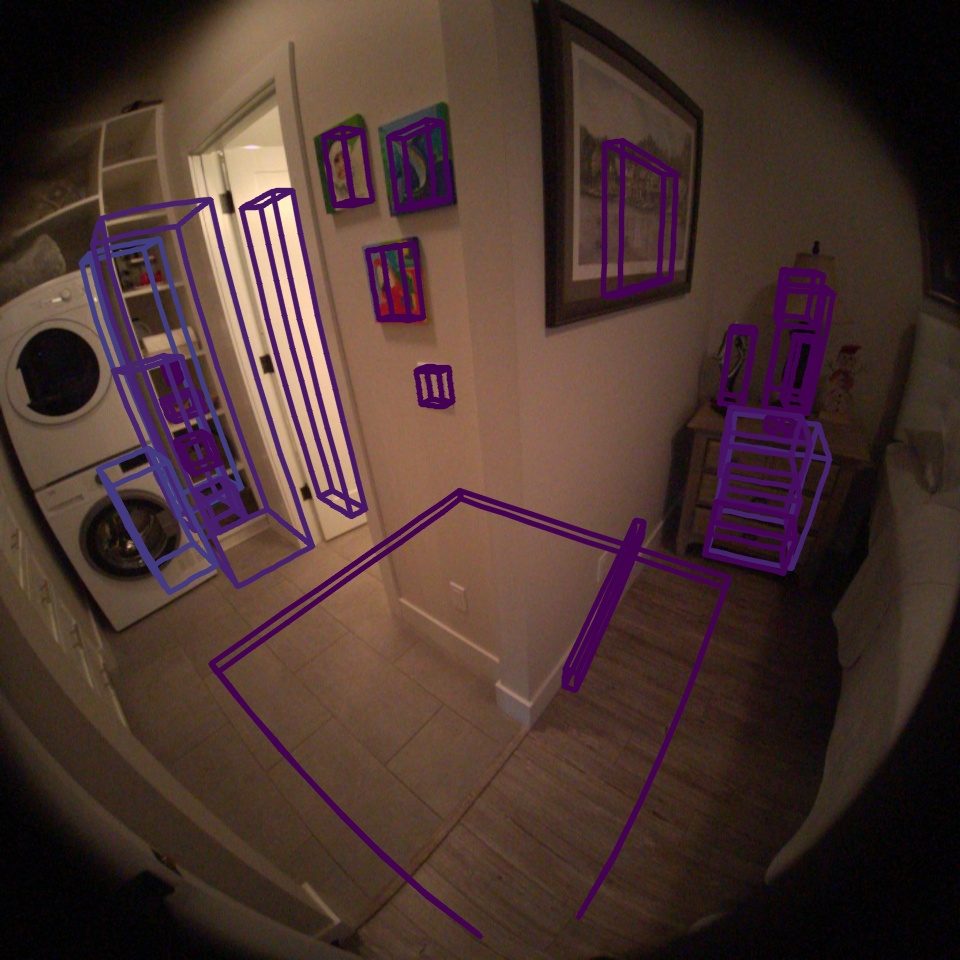}};
    \node(p03) at (p02.east) [anchor=west]{\includegraphics[width=\iw]{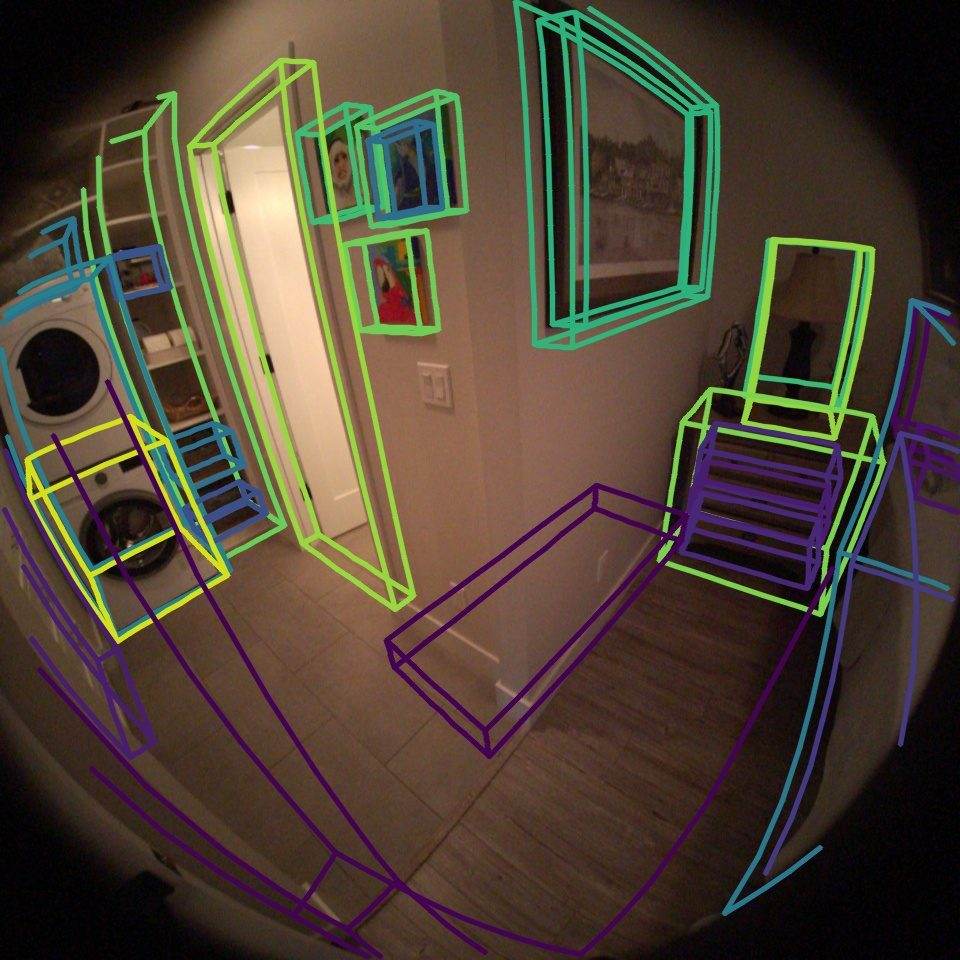}};
    \node(p04) at (p03.east) [anchor=west]{\includegraphics[width=\iw]{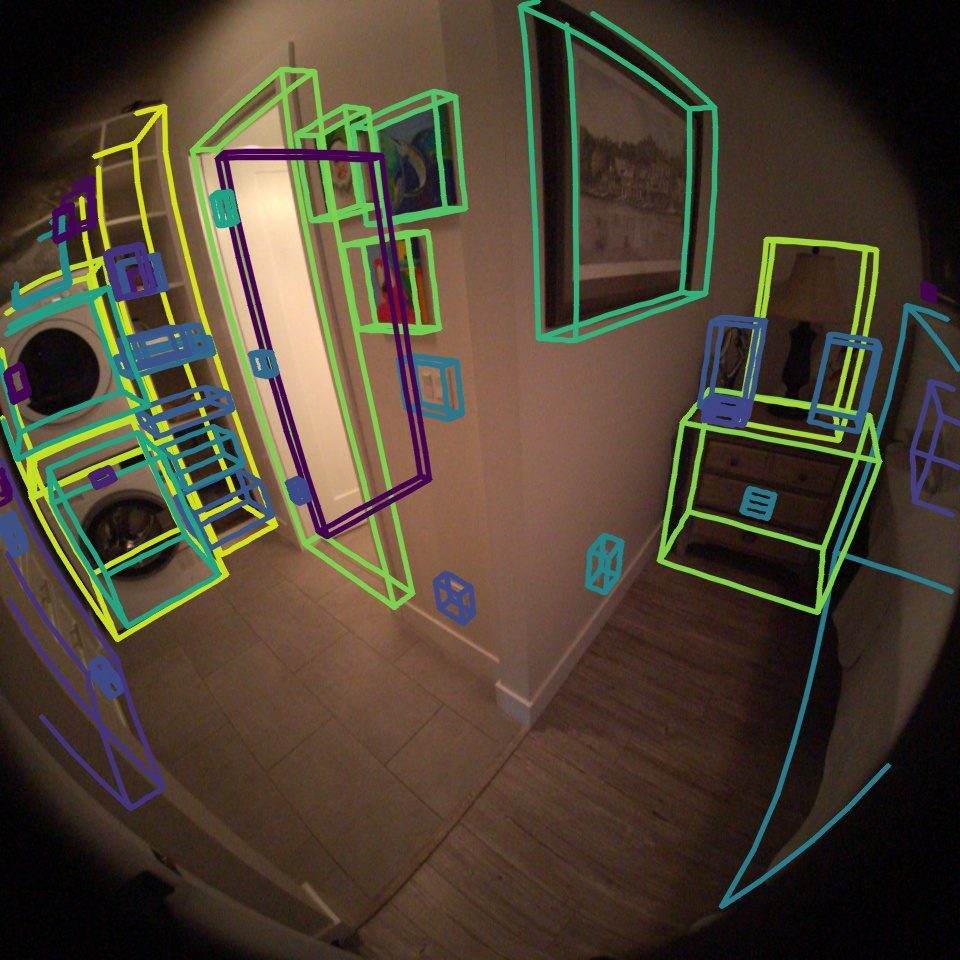}};

    \node(p11) at (p01.south) [anchor=north,yshift=\dy]{\includegraphics[width=\iw]{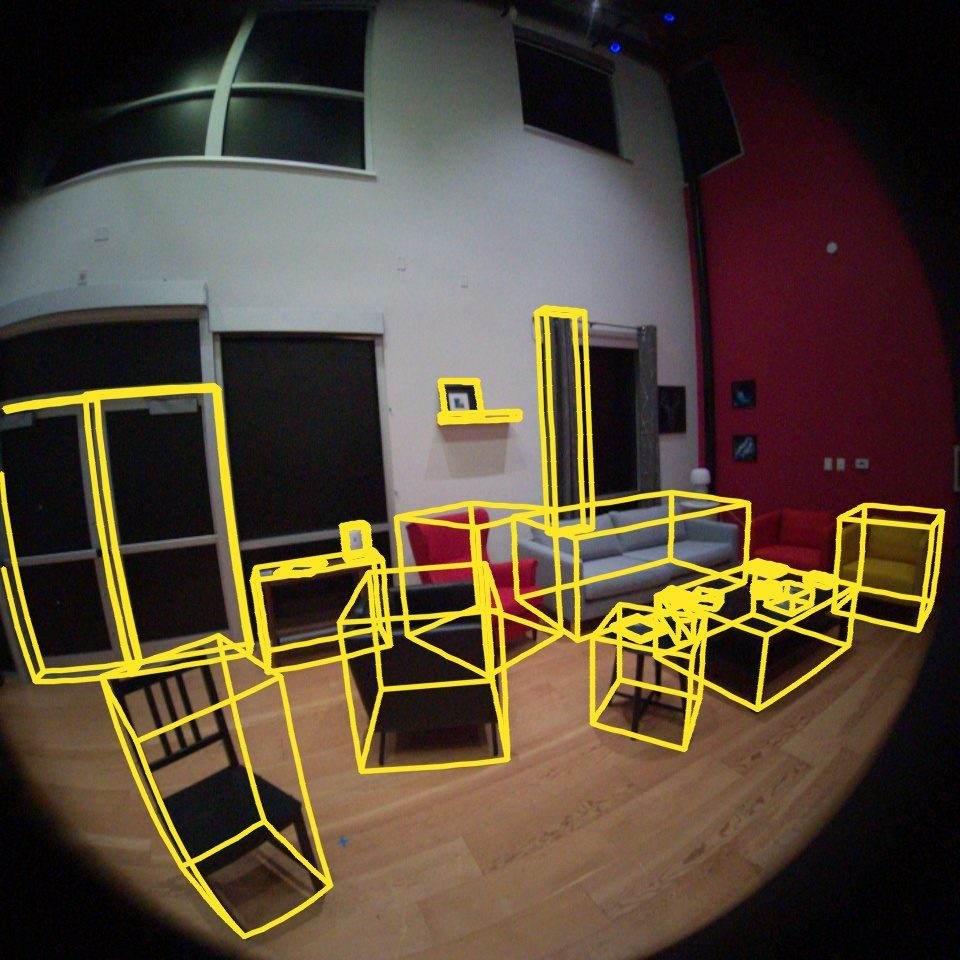}};
    \node(p12) at (p02.south) [anchor=north,yshift=\dy]{\includegraphics[width=\iw]{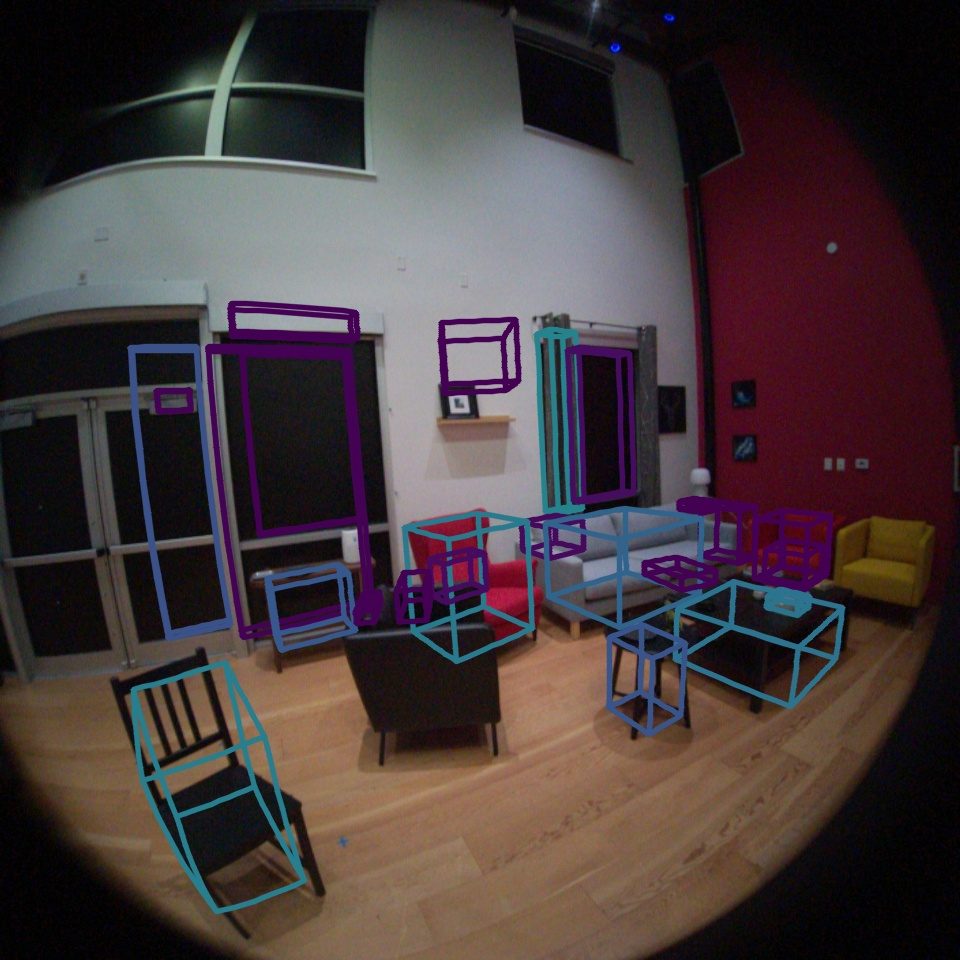}};
    \node(p13) at (p03.south) [anchor=north,yshift=\dy]{\includegraphics[width=\iw]{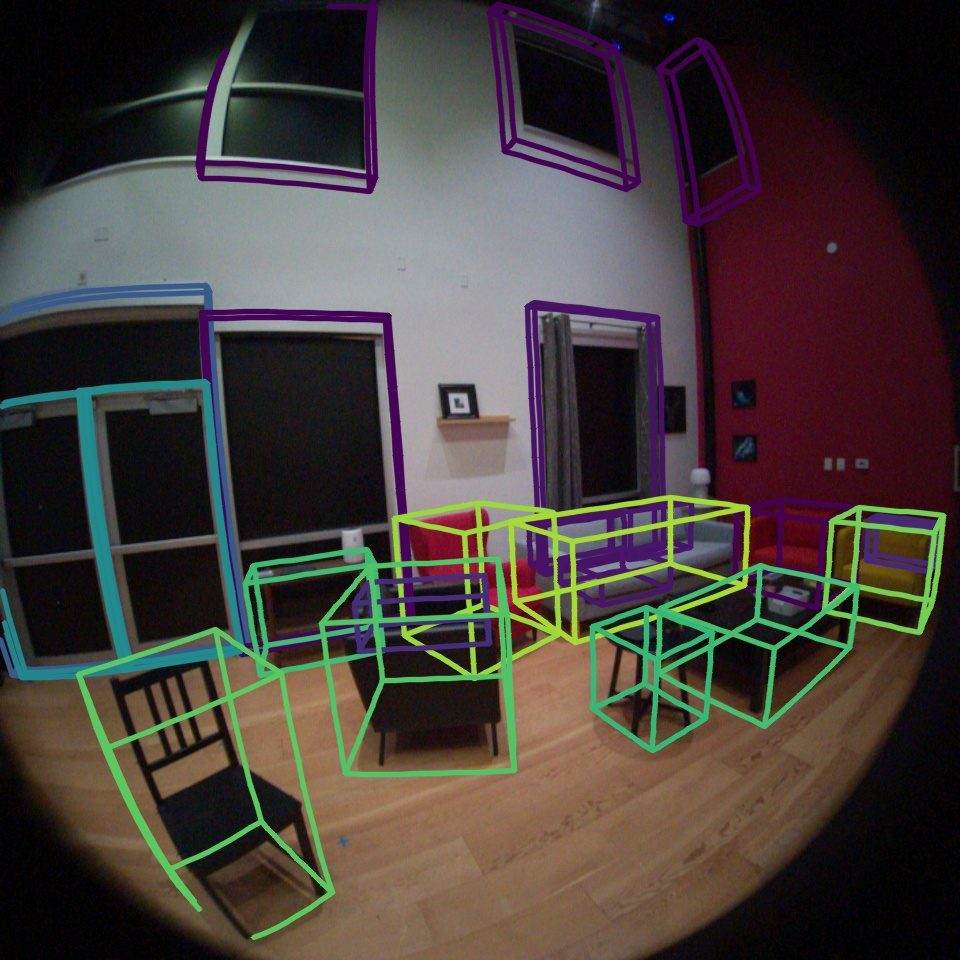}};
    \node(p14) at (p04.south) [anchor=north,yshift=\dy]{\includegraphics[width=\iw]{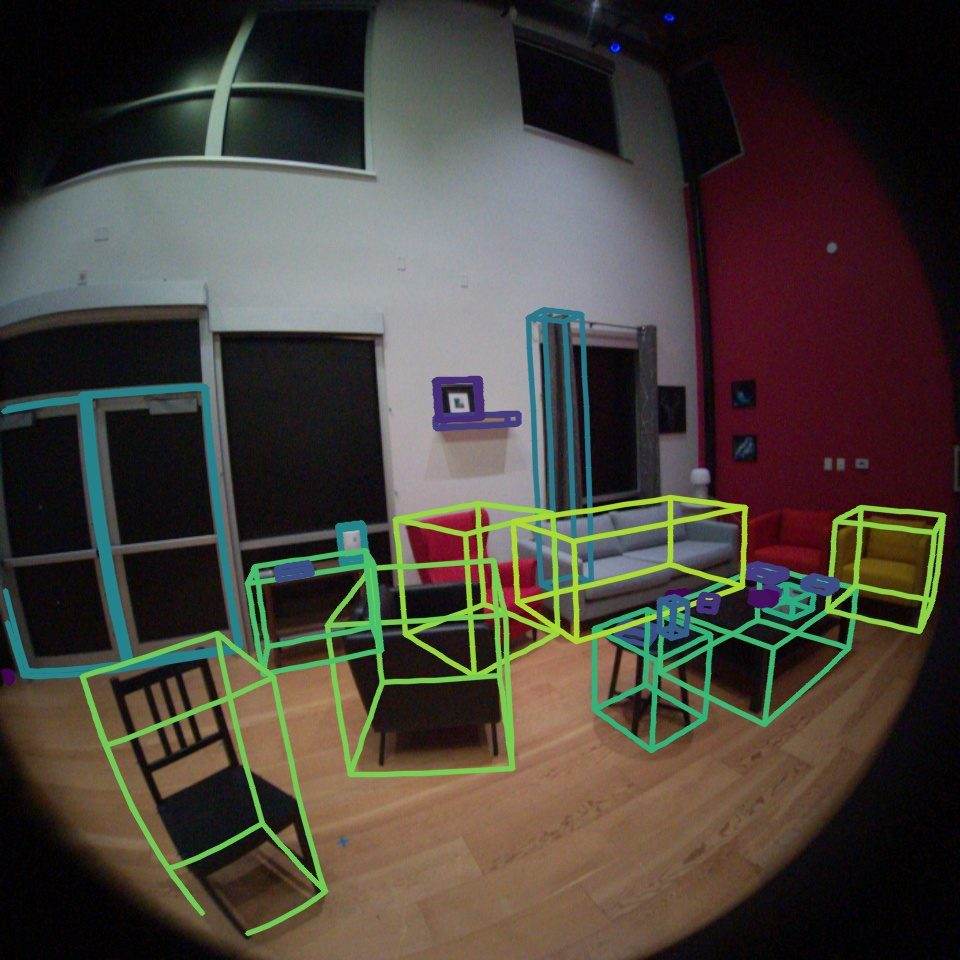}};

    \node(t0) at(p01.north) [anchor=south, text width=\iw, yshift=-\dy, align=center, font=\tiny]{Ground Truth};
    \node(t1) at(p02.north) [anchor=south, text width=\iw, yshift=-\dy, align=center, font=\tiny]{OWLv2+CuTR};
    \node(t2) at(p03.north) [anchor=south, text width=\iw, yshift=-\dy, align=center, font=\tiny]{OWLv2+BoxerNet(ours)};
    \node(t3) at(p04.north) [anchor=south, text width=\iw, yshift=-\dy, align=center, font=\tiny]{GT2D+BoxerNet(ours)};

    %%% For 3DMOOD

    \node(p31) at (p11.south) [anchor=north,yshift=-1em]{\includegraphics[width=\iw]{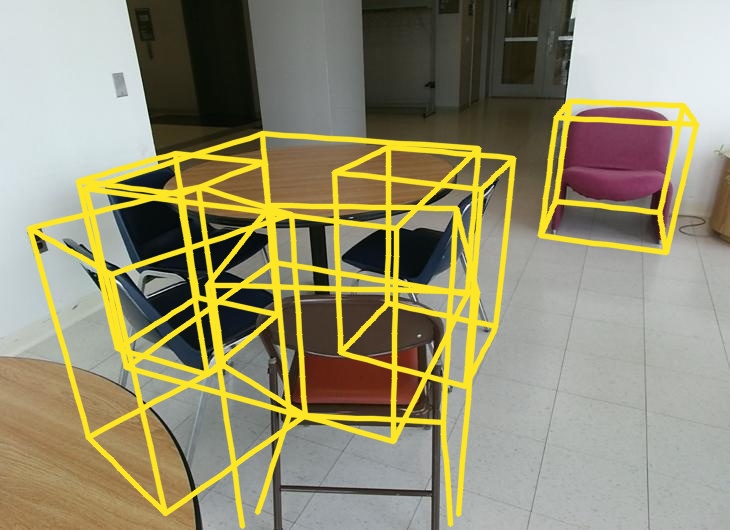}};
    \node(p32) at (p12.south) [anchor=north,yshift=-1em]{\includegraphics[width=\iw]{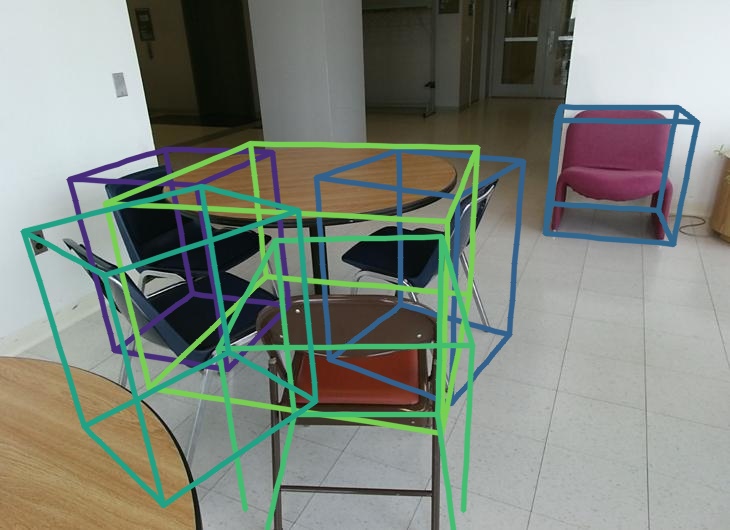}};
    \node(p33) at (p13.south) [anchor=north,yshift=-1em]{\includegraphics[width=\iw]{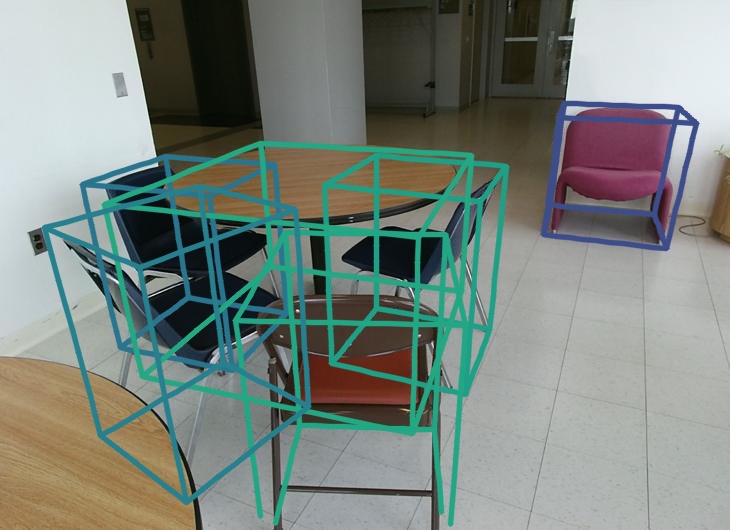}};
    \node(p34) at (p14.south) [anchor=north,yshift=-1em]{\includegraphics[width=\iw]{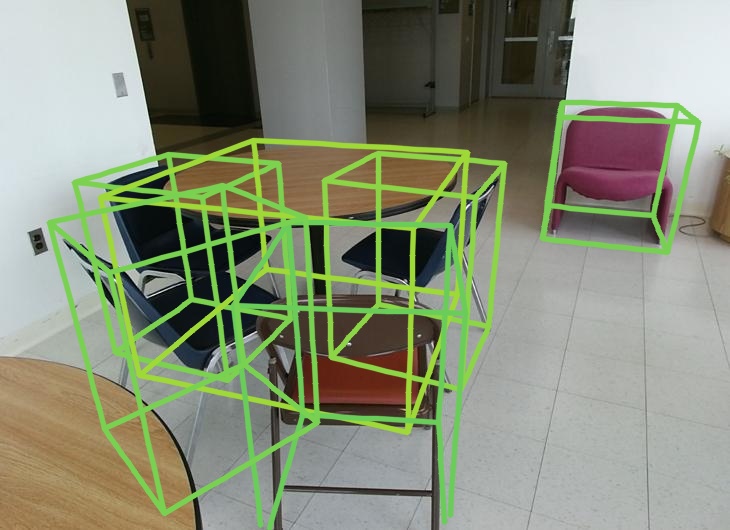}};

    \node(t00) at(p31.north) [anchor=south, text width=\iw, yshift=-\dy, align=center, font=\tiny]{Ground Truth};
    \node(t01) at(p32.north) [anchor=south, text width=\iw, yshift=-\dy, align=center, font=\tiny]{3DMOOD(RGB)};
    \node(t02) at(p33.north) [anchor=south, text width=\iw, yshift=-\dy, align=center, font=\tiny]{3DMOOD+BoxerNet(RGB)};
    \node(t03) at(p34.north) [anchor=south, text width=\iw, yshift=-\dy, align=center, font=\tiny]{3DMOOD+BoxerNet(RGBD)};
    %%% END 3D MOOD

    \node(l0) at (p01.west) [anchor=south, rotate=90, yshift=-\dy, font=\tiny]{NymeriaPlus };
    \node(l1) at (p11.west) [anchor=south, rotate=90, yshift=-\dy, font=\tiny]{ADT};
    \node(l2) at (p31.west) [anchor=south, rotate=90, yshift=-\dy, font=\tiny]{SUN-RGBD};

    \node(cb) at (p31.south west) [anchor=north west, xshift=7em, yshift=-2pt]{\includegraphics[width=0.7\textwidth]{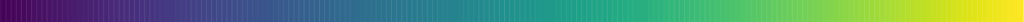}};
    \node(c1) at (cb.east) [anchor=west, font=\tiny]{1.0};
    \node(c2) at (cb.west) [anchor=east, font=\tiny]{IoU with Gt: \hspace{1ex} 0.0};

  \end{tikzpicture}
  \caption{\textbf{Per-frame 3D IoU visualization.} \textbf{First 2 rows:} We visualize the 3D IoU overlap of the ground truth left with OWLv2+CuTR (middle left), OWLV2+BoxerNet(middle right) and GT2D+BoxerNet(right). Predictions are colored in a viridis colormap (yellow is better). Boxer estimates have higher 3D IoU with the ground truth boxes as shown by the more yellow boxes. \textbf{Last row:} We also show the comparison of 3D-MOOD and BoxerNet using 3D-MOOD 2D boxes, on SUN-RGBD dataset.}
  \label{fig:qual-per-frame}
\end{figure}

\subsection{Per-Frame Experimental Analysis}
For all baseline methods, BoxerNet improves 3DBB estimation. Compared to CuTR, BoxerNet significantly outperforms it on NymeriaPlus in all configurations, even in image-only settings with ground-truth 2DBBs (0.296 vs. 0.010). One likely factor is that CuTR was not trained on egocentric data. Since NymeriaPlus only has sparse point-cloud depth, we do not run the CuTR RGB-D model and leave it as a dash ($-$). However, even on CA-1M, where both methods are trained and dense depth is available, we still see a gap (0.412 vs. 0.250). A qualitative visualization comparing CuTR and BoxerNet is shown in \cref{fig:qual-per-frame}. See the ablation study in \cref{sec:abl} for further analysis.

Compared to 3D-MOOD, BoxerNet also improves performance in both image-only and image+depth settings. BoxerNet uses a DINOv3 backbone, which has been shown to exhibit strong monocular depth priors across large datasets~\cite{simeoni2025dinov3}. This may help explain why it outperforms 3D-MOOD, which explicitly learns an auxiliary depth map on a smaller dataset.

% \TODO{add more explaination once results finalize}.

% \lingni{There should be another figure, just for per-frame 3D prediction. It should contain at least 4 columns, gt 3DBB rendered on an image; CUTR 3DBB on image, Boxer 2DBB on image using the best 2D detector, then Boxer 2DBB on imge using the gt 2d detector. We need to show the representative examples for each data sets. This should be the main quantitative figure for BoxerNet.}

% \lingni{Fig 4 and 5 can be merged into one figure, because they are both about heatmaps, just on different dataset. Per row, I will show 5 columns: 1-2 reference Image views, gt 3dBB, CUTR result, boxer result using the best 2D detector, boxer result using the gt detector. I'll make the figure on white background if possible, but bg color is not top priority. This should be the main quantitative figure for Boxer algorithm with multiview fusion.}

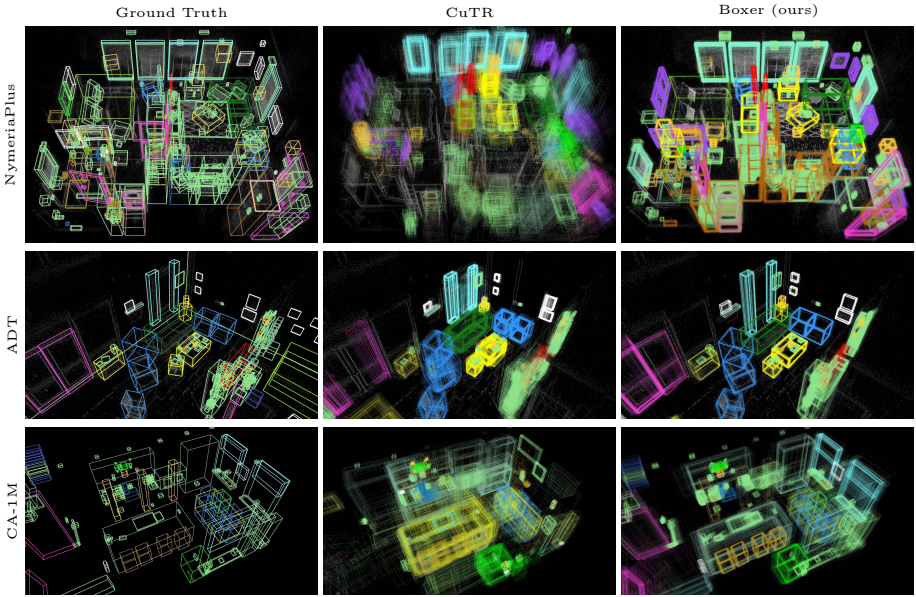
\begin{figure}[t]
  \centering
  \def\iw{12em}
  \def\dy{-1pt}
  \begin{tikzpicture}[inner sep=1pt]

    % --- Top row ---
    \node(p00)                            {\includegraphics[width=\iw]{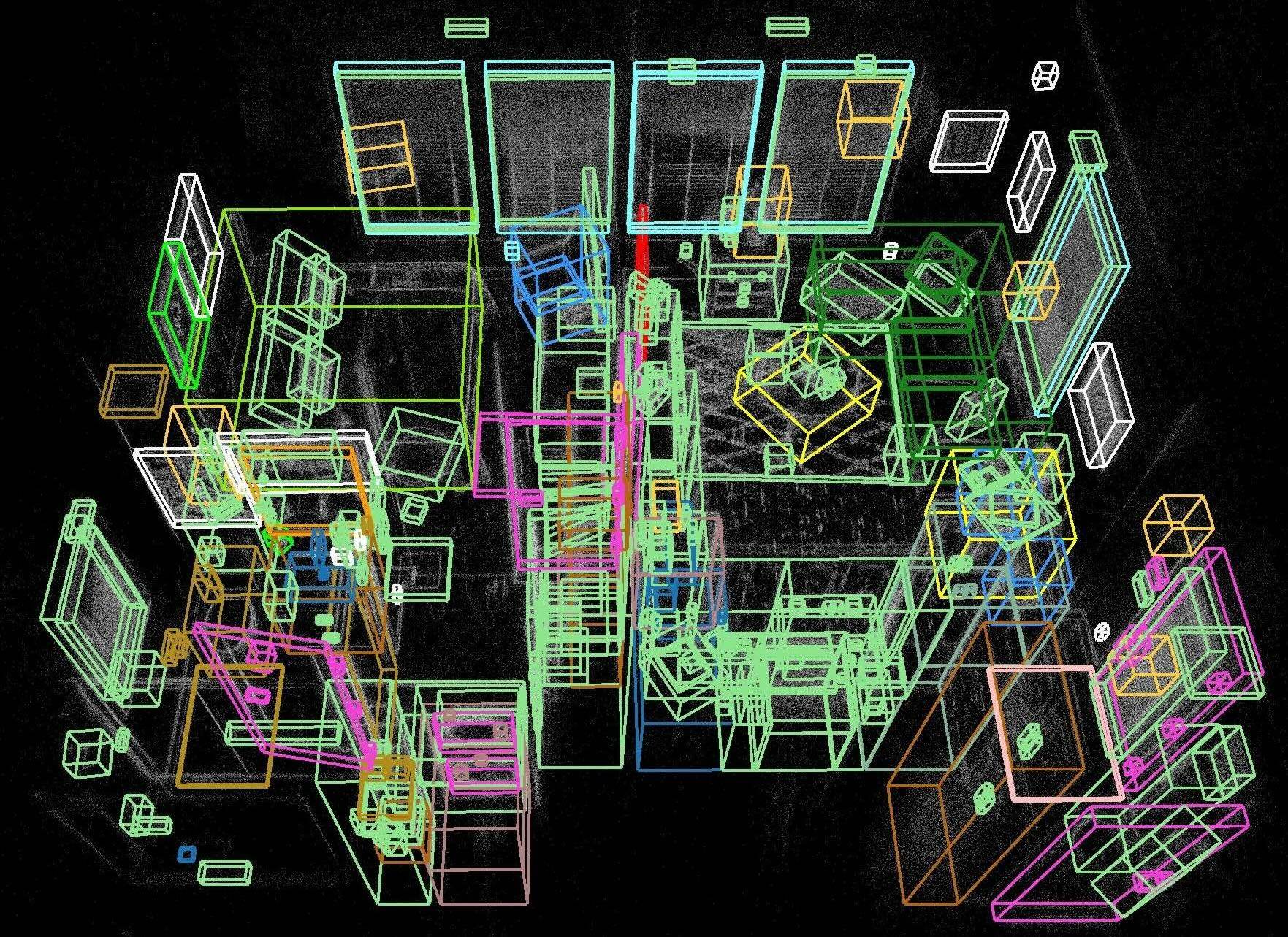}};
    \node(p01) at (p00.east) [anchor=west]{\includegraphics[width=\iw]{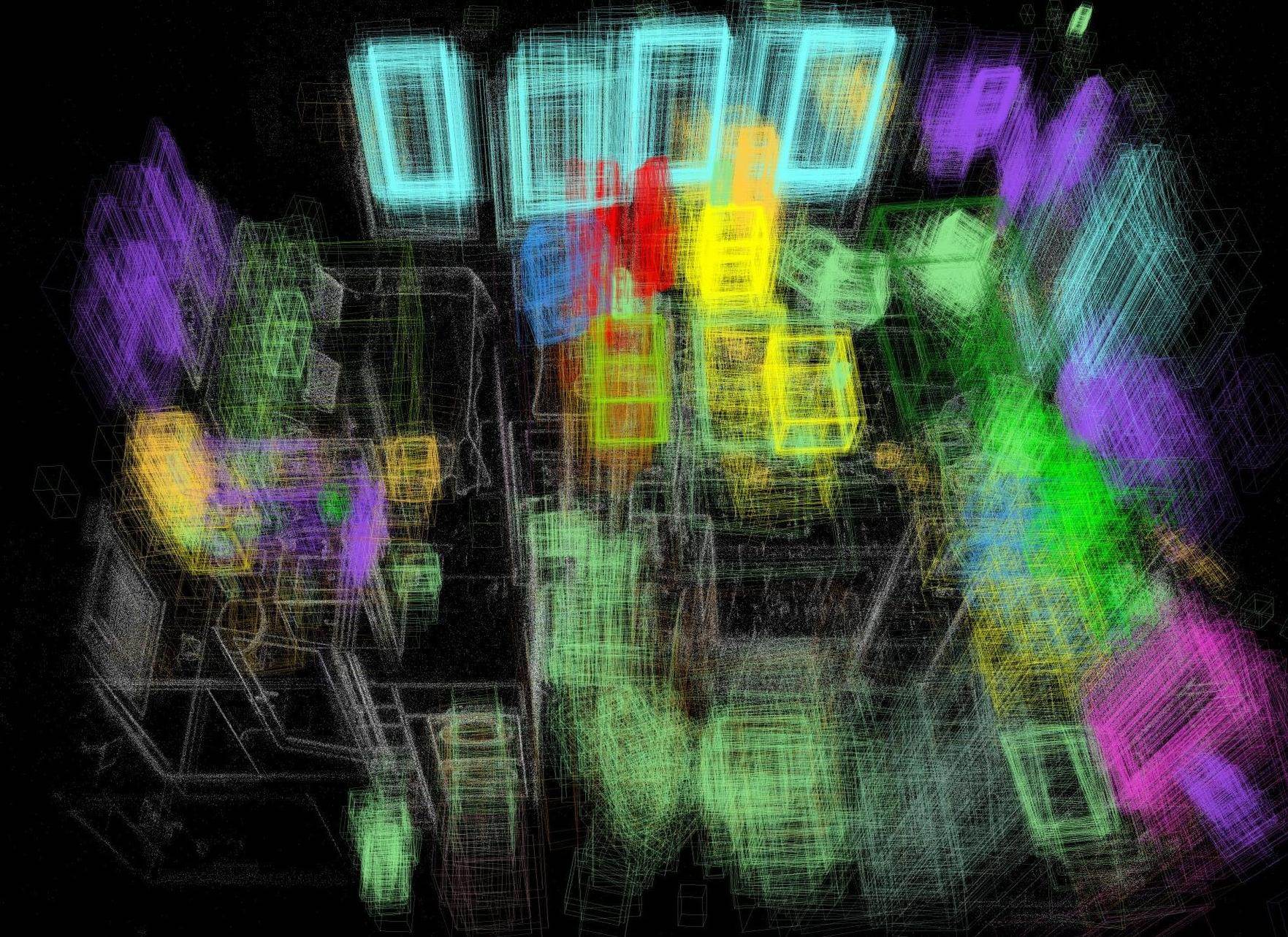}};
    \node(p02) at (p01.east) [anchor=west]{\includegraphics[width=\iw]{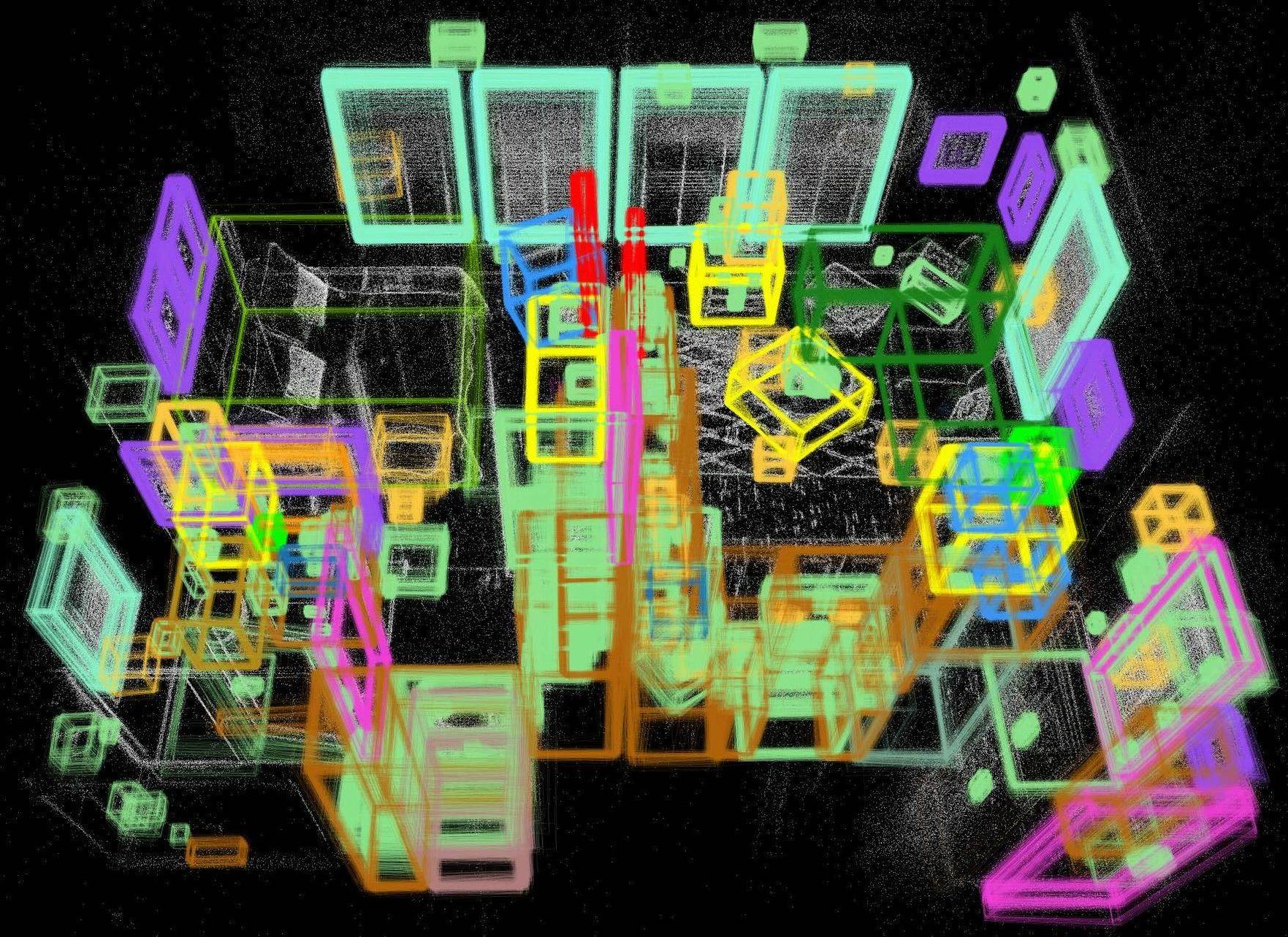}};

    % --- Second row ---
    \node(p10) at (p00.south) [anchor=north,yshift=\dy]{\includegraphics[width=\iw]{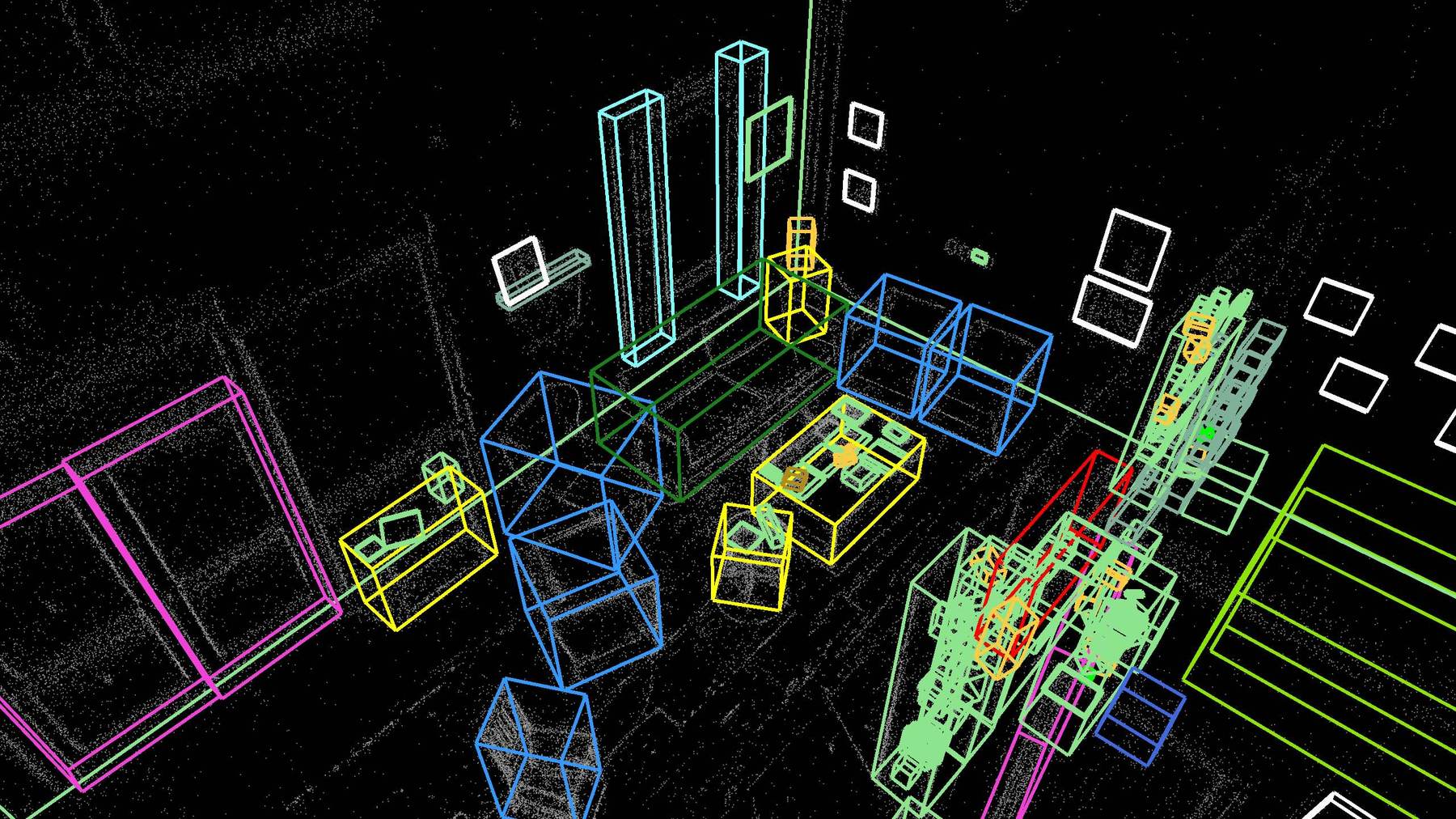}};
    \node(p11) at (p10.east) [anchor=west]{\includegraphics[width=\iw]{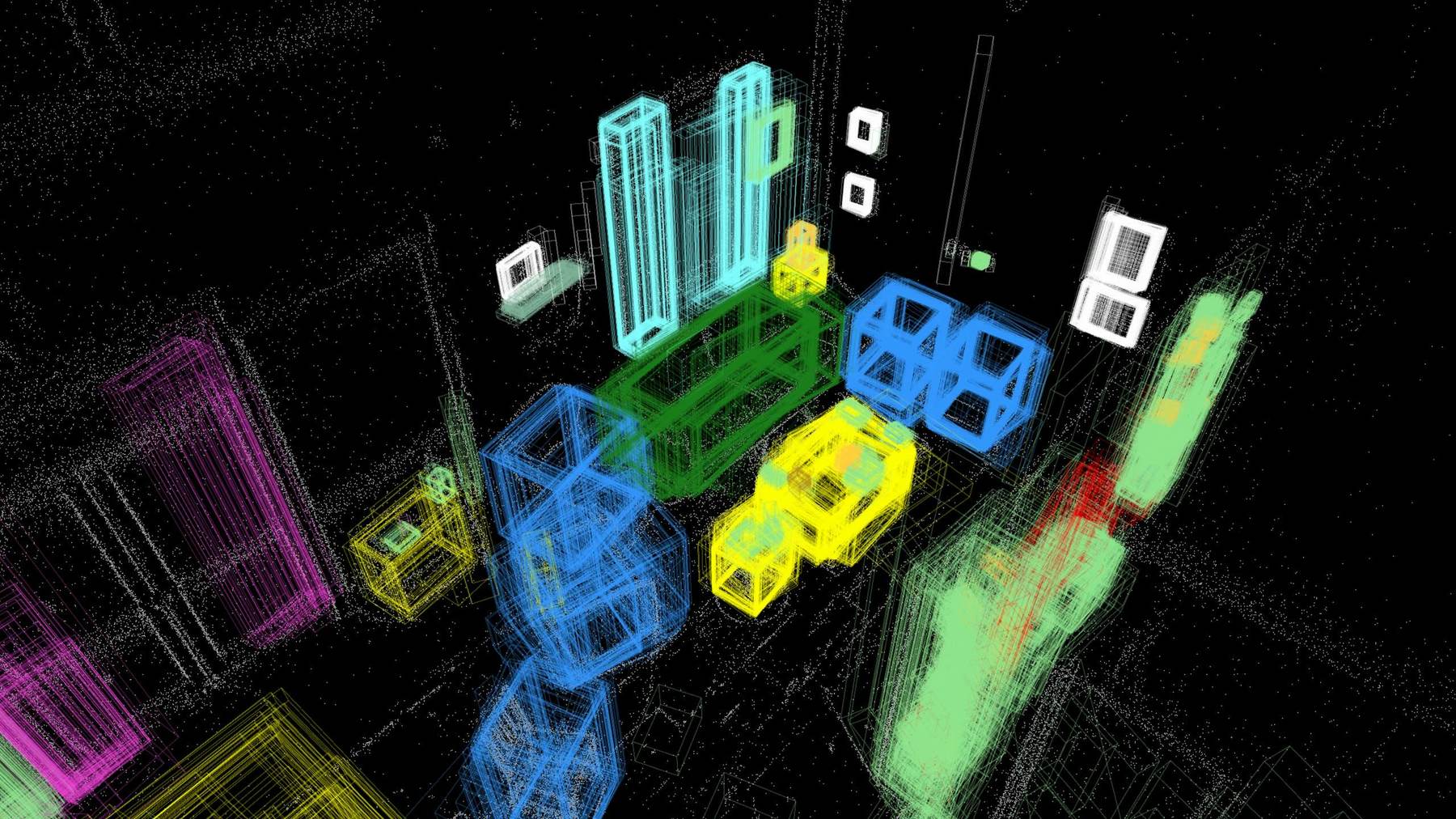}};
    \node(p12) at (p11.east) [anchor=west]{\includegraphics[width=\iw]{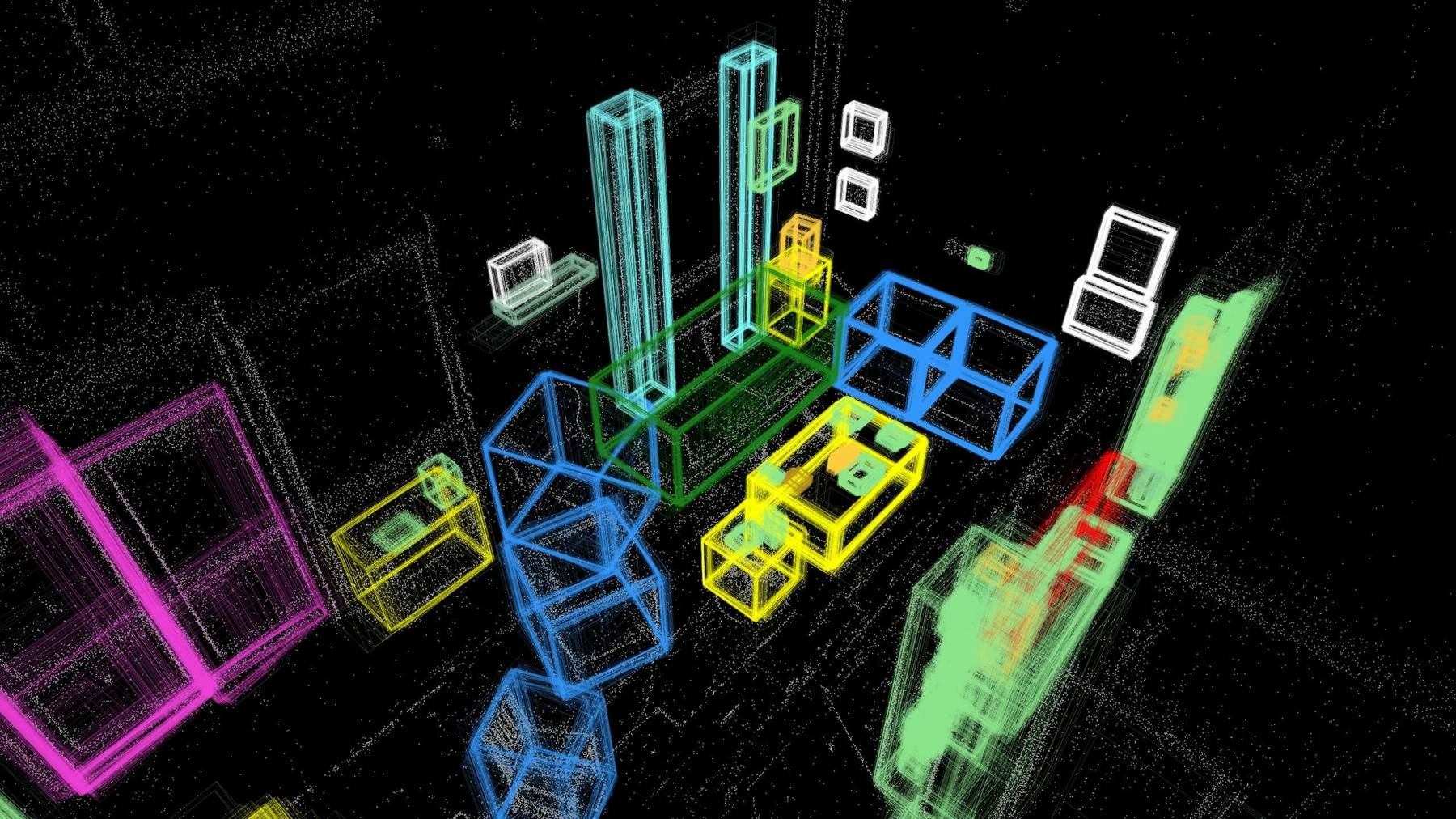}};

    % --- Third row ---
    \node(p20) at (p10.south) [anchor=north,yshift=\dy]{\includegraphics[width=\iw]{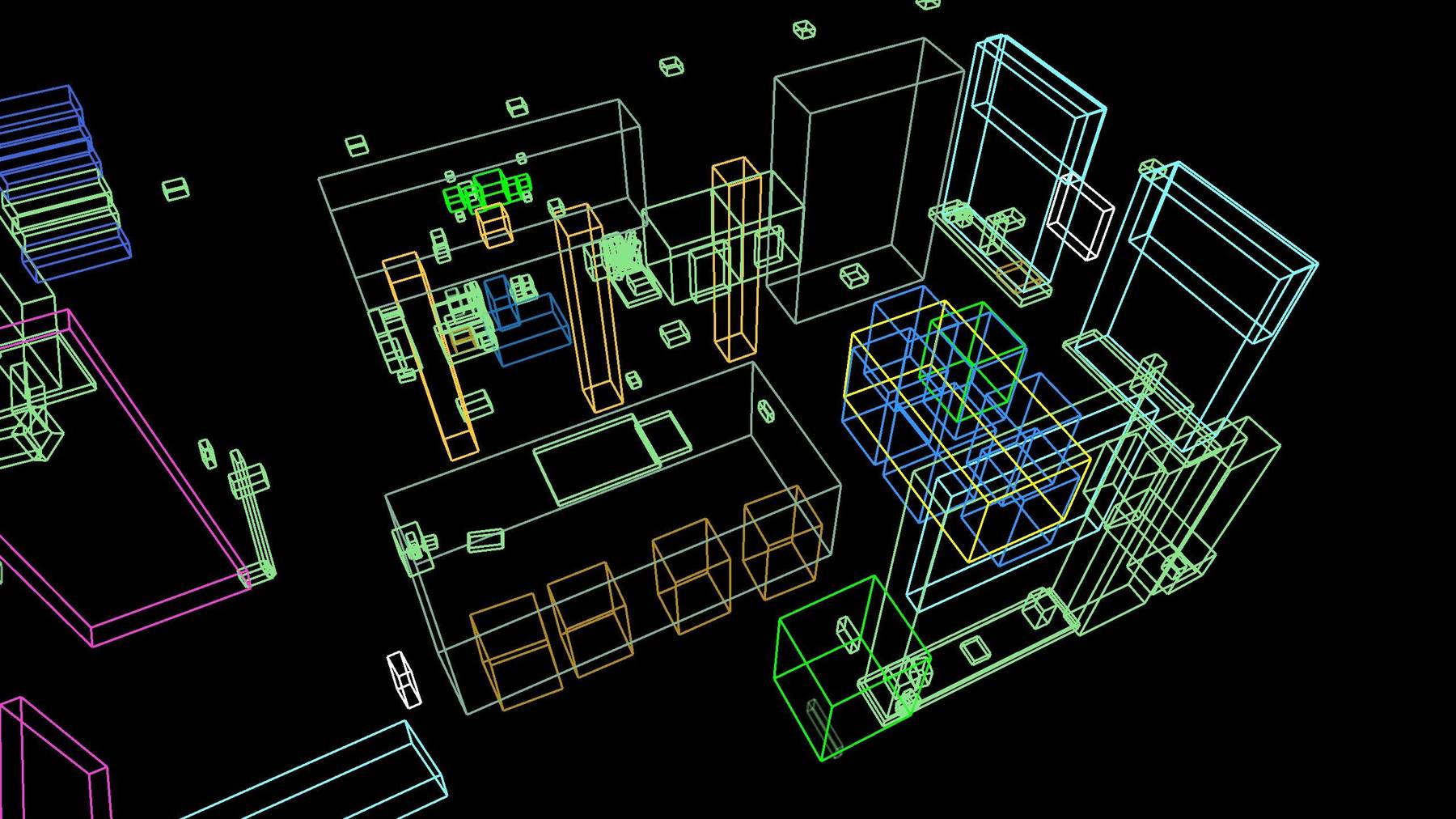}};
    \node(p21) at (p20.east) [anchor=west]{\includegraphics[width=\iw]{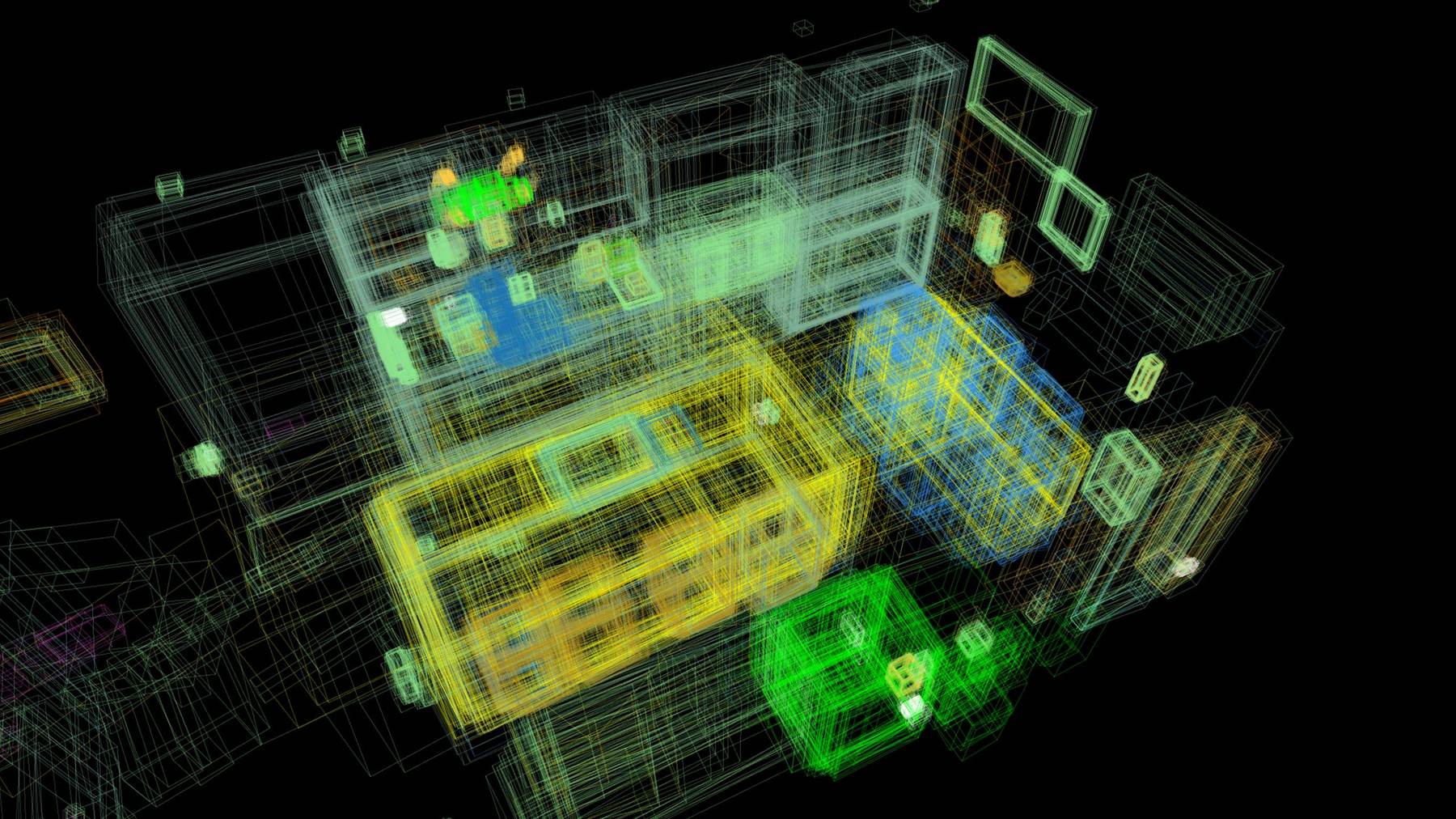}};
    \node(p22) at (p21.east) [anchor=west]{\includegraphics[width=\iw]{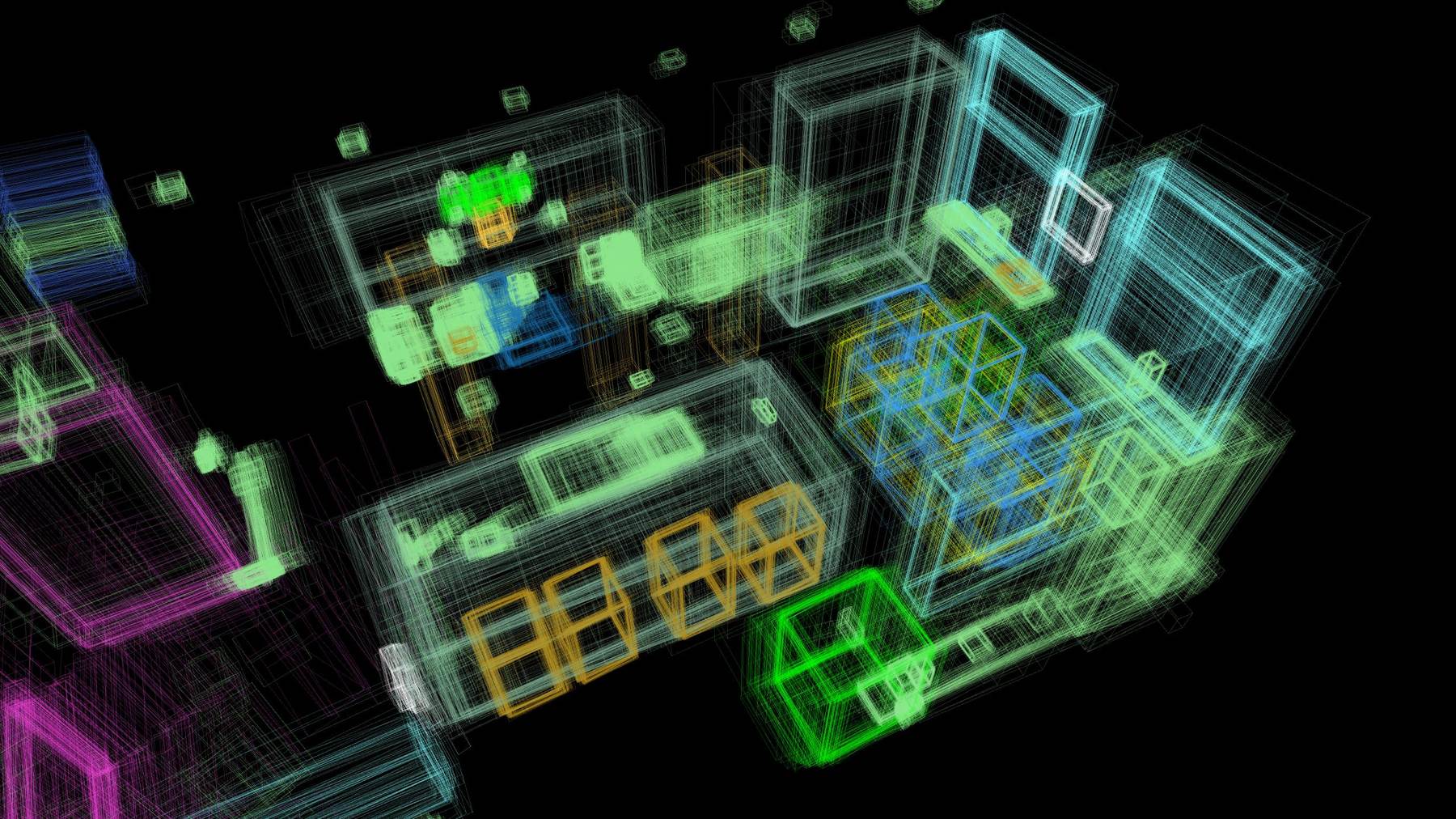}};

    % --- Column titles ---
    \node(t0) at(p00.north) [anchor=south, text width=\iw, yshift=-\dy, align=center, font=\tiny]{Ground Truth};
    \node(t1) at(p01.north) [anchor=south, text width=\iw, yshift=-\dy, align=center, font=\tiny]{CuTR};
    \node(t2) at(p02.north) [anchor=south, text width=\iw, yshift=-\dy, align=center, font=\tiny]{Boxer (ours)};

    % --- Row labels ---
    \node(l0) at (p00.west) [anchor=south, rotate=90, yshift=-\dy, font=\tiny]{NymeriaPlus};
    \node(l1) at (p10.west) [anchor=south, rotate=90, yshift=-\dy, font=\tiny]{ADT};
    \node(l2) at (p20.west) [anchor=south, rotate=90, yshift=-\dy, font=\tiny]{CA-1M};

  \end{tikzpicture}
  \caption{\textbf{Lifted per-frame 3DBB pseudo heatmaps.} We compare the Ground Truth per-scene 3DBBs (left) to all the per-frame 3DBBs from CuTR (middle) and Boxer (right), prompted with GT 2DBB input, into a consistent coordinate frame and show the boxes rendered on top of one another creating a pseudo-heatmap. Boxer exhibits a sharper heatmap compared to CuTR which corresponds to more consistent predictions. Colors loosely correspond to object semantic class.}
  \label{fig:qual-per-scene}
\end{figure}

% \begin{figure}[!ht]
%   \centering
%   \includegraphics[width=0.98\textwidth]{figures/misc/qualheatv2.jpg}
%   \caption{\textbf{3D Detection Heatmaps on CA-1M.}
%   Per-frame 3D predictions of various techniques are visualized in a shared coordinate from (pre-fusion) in a heatmap style visualization on the CA-1M validation set. \TODO{add more methods. add Image viz as well}\lingni{add the meaning of colors.}}
%   \label{fig:qual}
% \end{figure}

% \begin{figure}[!ht]
%   \centering
%   \includegraphics[width=0.98\textwidth]{figures/misc/qualadtv1.jpg}
%   \caption{\textbf{3D Detection Heatmaps on ADT.}
%   Per-frame 3D predictions of various techniques are visualized in a shared coordinate from (pre-fusion) in a heatmap style visualization on the ADT dataset. \TODO{add more methods. add Image viz as well}}
%   \label{fig:qual_adt}
% \end{figure}

\subsection{Per-Scene Experimental Analysis} We focus on video datasets (excluding SUN-RGBD) and compare primarily to CuTR, since it is the closest baseline to BoxerNet and runs in real time. As shown in \cref{tab:results_scene}, the trend is consistent with the per-frame results: BoxerNet improves performance in all cases. \cref{fig:qual-per-scene} compares the per-frame predictions between CuTR and Boxer lifted into a consistent global map, demonstrating that Boxer exhibits more consistent predictions between frames.

\subsection{Ablation and Sensitivity Studies}
\label{sec:abl}

We study key design decisions in the BoxerNet training recipe. We evaluate the 2D-to-3D lifting capability of the model on both egocentric and non-egocentric data sources. Ground-truth (oracle) 2D bounding boxes are used as input to best isolate the performance of the 3D lifting model.

\begin{table}[ht]
  \centering
  \tighttable
  \scriptsize
  \setlength{\tabcolsep}{4pt}
  \renewcommand{\arraystretch}{1.05}
  \caption{\textbf{Ablation study for BoxerNet.} We compare removing individual design components using GT 2DBB inputs and show per-frame mAP $\uparrow$. \vspace{-1ex}}
  \begin{tabular}{l c c}
    \toprule
    Variant                                     & NymeriaPlus          & CA-1M           \\
    \midrule
    Full BoxerNet                                 & \textbf{0.518}   & \textbf{0.412}  \\
    \hline
    w/o data augmentation                       & 0.497            & 0.400           \\
    w/o aleatoric uncertainty head              & 0.485            & 0.401           \\
    half image resolution (480x480 vs 960x960)    & 0.466            & 0.395           \\
    only public training data                   & 0.463            & 0.376           \\
    w/o depth                                   & 0.279            & 0.126           \\
    only CA-1M                                  & 0.002            & 0.357           \\

    \bottomrule
  \end{tabular}
  \label{tab:boxer_ablation}
\end{table}

% \fan{curious any insights why no gains on data augmentation+uncertainy for CA-1M}\vspace{-1ex}

\myparagraph{Analysis of \cref{tab:boxer_ablation}.} We ablate different design choices in BoxerNet. Removing point-cloud input has a significant effect, reducing performance from 0.518 to 0.279 mAP on the NymeriaPlus test set. Removing aleatoric uncertainty also hurts performance, from 0.518 to 0.485. There is a large domain gap between CA-1M and Aria data, since performance drops from 0.518 to 0.002 when training only on CA-1M. Since BoxerNet is trained partly on internal data, we also measure the impact of removing internal data from training: performance drops from 0.518 to 0.463 on NymeriaPlus and from 0.412 to 0.376 on CA-1M.

\subsection{Limitations}

BoxerNet was trained on static objects, and the fusion system assumes a static world, so dynamic objects (\textit{e.g.,} in-hand objects) do not work very well. Highly non-cuboidal objects (\textit{e.g.,} wires, plant vines) are not well represented by a 3DBB, and the model is not expected to perform well in these circumstances. BoxerNet lifting requires calibrated input and gravity. If not available they could be acquired from off-the-shelf methods such as ~\cite{veicht2024geocalib}, which we leave to future work to explore. OWLv2 and DETIC make mistakes and are not trained heavily on egocentric data. Boxer inherits these limitations since it is based on these open-set 2D detectors.

\section{Conclusion}
\label{sect:conclusion}

We presented Boxer, an algorithm for generating global, static, and de-duplicated 3D object bounding boxes from posed video sequences with optional depth. By combining an open-vocabulary 2D detector with BoxerNet, a learned 2D-to-3D lifting model trained on a large dataset consisting of multiple camera types, Boxer produces stable 3D object bounding boxes for open-world scenes. BoxerNet itself can be used as a drop-in replacement to existing 2D or 3D object detection pipelines to improve 3DBB accuracy. A multi-view temporal fusion module integrates per-frame predictions into a unified global map, using 3D IoU and heuristic rules to merge redundant instances and refine object estimates. Overall, Boxer provides a practical and scalable approach for constructing consistent scene-level 3D object representations from posed image data.

\appendix
\clearpage
\section*{Supplementary Material}
\addcontentsline{toc}{section}{Supplementary Material}
\section{Additional Boxer Algorithm Details}

In this section, we provide additional details on the Boxer algorithm.

\subsection{Offline Multi-View Fusion Details}
\label{sec:offline}

We fuse per-frame 3D detections into a consistent set of scene-level object hypotheses using a multi-stage geometric and semantic aggregation pipeline.
Let
\[
  \mathcal{B}^{3D} = \{ (b_i^{3D}, c_i, s_i^{3D}) \}_{i=1}^N
\]
denote the set of lifted 3D bounding boxes across frames, where $b_i^{3D}$ is a 7-DoF 3D box,
$c_i$ is the associated semantic label or text prompt, and $s_i^{3D} \in [0,1]$ is a confidence score. We merge detections that are geometrically overlapping, semantically consistent, and temporally redundant.
% \lingni{I think we used $b^{3D}$ to refer bounding box in main paper}

\subsubsection{3D IoU Filtering.}
We first compute the pairwise 3D intersection-over-union (IoU)
$\operatorname{IoU}_{3D}(b_j^{3D}, b_k^{3D})$ among all detected 3D bounding boxes.
Two detections are considered geometrically compatible if
\[
  \operatorname{IoU}_{3D}(b_j^{3D}, b_k^{3D}) \ge \tau_{\text{iou}} \;\;.
\]
This step restricts fusion to boxes that occupy overlapping volumes in 3D space and prevents merging detections that are spatially distant. By default, we use $\tau_{\text{iou}} = 0.3$.

% \lingni{have we described the threshold used for experiments somewhere?}

\subsubsection{Semantic Filtering.}
To avoid fusing detections of different categories, we apply semantic filtering prior to aggregation. Only detections with sufficiently similar semantic labels, as determined by a text embedding \cite{sbert}, are eligible for merging. This ensures that geometrically overlapping but semantically distinct objects are handled separately.

% \js{This hard constraint setup is suitable for the large yet still finite vocabularies used to prompt the 2DBB detectors. This hard constraint could be relaxed by using word-embedding distanes for example and thus allow truly open-world semantic filtering.}
% dd -- that was outdated, we actually use a text embedding for this, good catch.

\subsubsection{Clustering via Connected Components.}
We construct an undirected graph $G=(V,E)$ whose nodes $v_i \in V$ correspond to detections $b_i^{3D}$.
An edge $(j,k) \in E$ exists if both the geometric and semantic criteria are satisfied.
Connected components analysis (CCA) on $G$ yields a set of clusters
$\{ \mathcal{C}_m \}_{m=1}^M$,
where each cluster is hypothesized to correspond to a single physical object observed across time.

\subsubsection{Rotation-Aware Averaging.}
Within each cluster $\mathcal{C}_m$, we fuse detections into a single 3D bounding box using
confidence-weighted averaging.
To account for the $90^\circ$ rotational symmetry of gravity-aligned boxes, we first align all
detections to a canonical orientation by selecting, for each box, the representation (original or
$90^\circ$-rotated with swapped in-plane dimensions) that best agrees with the cluster consensus.
After alignment, object position and size are averaged, and the final yaw angle is computed using
a circular mean. The rotation and the scales are swapped accordingly in such cases.

\subsubsection{Non-Maximum Suppression.}
Finally, we apply 3D non-maximum suppression (NMS) to the fused boxes.
Given fused hypotheses $\{ \tilde{b}_m^{3D} \}$, boxes with
\[
  \operatorname{IoU}_{3D}(\tilde{b}_m^{3D}, \tilde{b}_n^{3D}) \ge \tau_{\text{nms}}
\]
are suppressed in favor of the higher-confidence prediction.
This produces a compact set of final scene-level 3D object detections. By default, we use $\tau_{\text{nms}} = 0.6$.

\subsection{Online Multi-View Tracker Details}
For all experiments in the paper, we use the offline tracker to generate per-scene 3DBBs. One limitation of this approach is that it scales as $O(N^2)$, where $N$ is the total number of per-frame estimated 3DBBs in a sequence. In practice, runtime remains reasonable with an efficient 7-DoF IoU algorithm (e.g., a 10-minute sequence processes in under 30 seconds on a Lenovo P620 workstation with an RTX 4090 GPU), though most tracker operations other than IoU computation run on CPU.

To scale to longer sequences and support streaming detection, we enable Boxer's online tracking mode. This mode generates a set of $M$ 3DBB hypotheses once an instance has been observed sufficiently often. For each incoming set of $M$ per-frame 3DBBs, hypotheses are matched to the set of visible tracked boxes $P$ (as determined by observed depth). This yields a complexity of $O(MP)$, where both $M \ll N$ and $P \ll N$, and enables scaling to much longer sequences. We follow a similar implementation to \cite{straub2024efm3d}. %, which has more details and open-source code available. \TODO{add a quick comparison}

\subsection{BoxerNet Architecture Details.}

In our model, we use 4 self-attention layers and 6 cross-attention layers with 768 dimensions and 12 heads. We use DINOv3 Base. The two output MLPs each have two layers with 128 hidden units. In total BoxerNet has ~71 million parameters (excluding DINOv3).

\subsection{Rationale to Combine 2D and 3D Confidence Scores in Boxer}
% \lingni{I'd name it "rational/motivation to combine 2D and 3D confidence"}
% \js{Why and how ?}

Boxer's ability to rank detections by confidence has a large effect on mean Average Precision (mAP). As described in the main paper, we use the mean of the 2DBB and 3DBB detection scores as the final ranking score:

\begin{equation}
  s_i = (s_i^{2D} + s_i^{3D}) / 2 \;\;
\end{equation}

% \lingni{in previous text, you used $\sigma$ for confidence}
% dd -- added clarification, network outputs logvar sigma

The model predicts a log-variance $\sigma$, which is converted into a confidence score $s \in [0,1]$ as:

\[
s = \frac{1}{1 + e^{\sigma}} = \operatorname{sigmoid}(-\sigma)
\]

\begin{figure}[!htbp]
  \centering
  \includegraphics[width=0.6\textwidth]{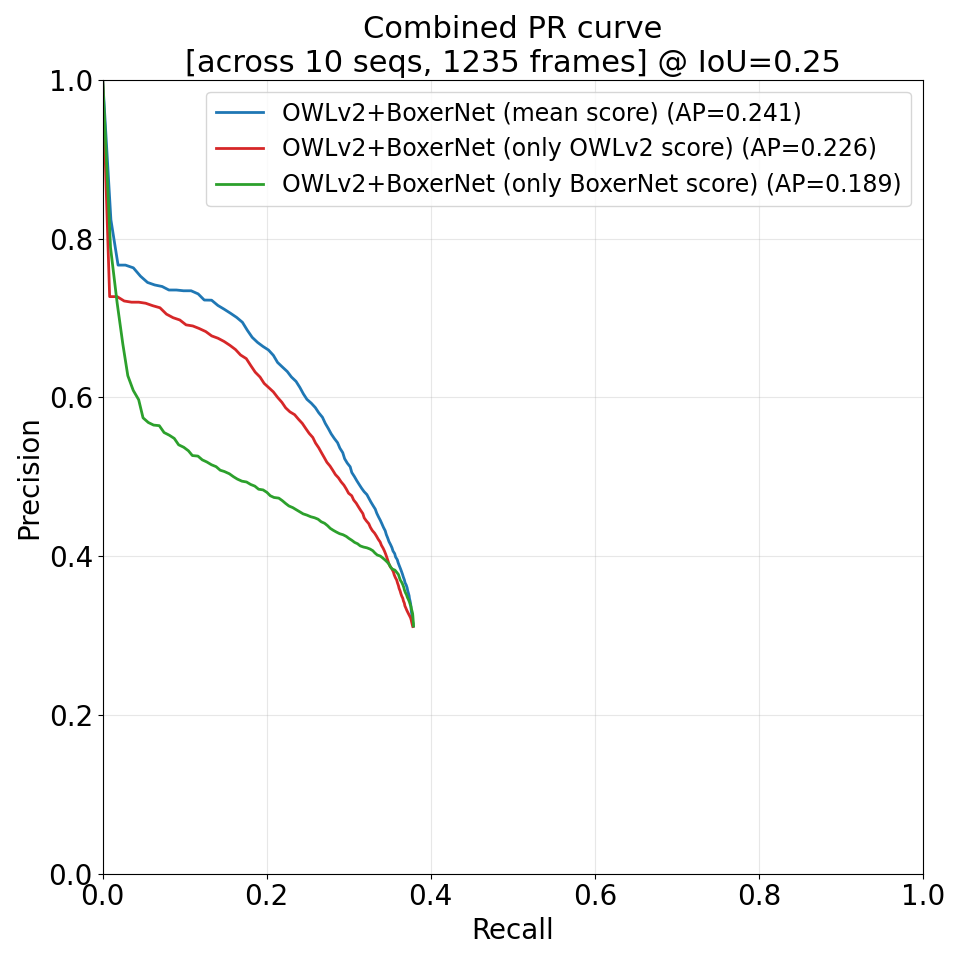}
  \caption{\textbf{Importance of Aleotoric Uncertainty for Ranking.}
    The Precision-Recall (PR) curve is shown for IoU=0.25 for three variants of BoxerNet, highlighting the importance of a good 3DBB scoring function enabled by using the mean of both 2D and 3D detection confidence scores.
  }
  \label{fig:combo_pr}
\end{figure}

On CA-1M, we compare three baselines: using only $s_i^{2D}$, using only $s_i^{3D}$, and using their average. As shown in \cref{fig:combo_pr}, averaging performs best. OWLv2-only (red) corresponds to $s_i^{2D}$, BoxerNet-only (green) corresponds to $s_i^{3D}$, and the averaged score (blue) gives the strongest performance.

%  \lingni{adding some visual examples might help builds the ituition that these two scores are often complementary to each other.}
% \js{I really liked the samples you showed in the meeting to give intuition as to why 2D and 3D uncertainty are complimentary.}
% \js{if you have time try the combination used in CubeRCNN to anchor this.}

% This can be explained intuitively through some examples.

\subsection{Data Augmentation Details.}

\begin{figure}[!htbp]
  \centering
  \def\iw{9em}
  \def\dy{-1pt}
  \begin{tikzpicture}[inner sep=1pt]
    % Row 1: Photometric
    \node(p11)                            {\includegraphics[width=\iw]{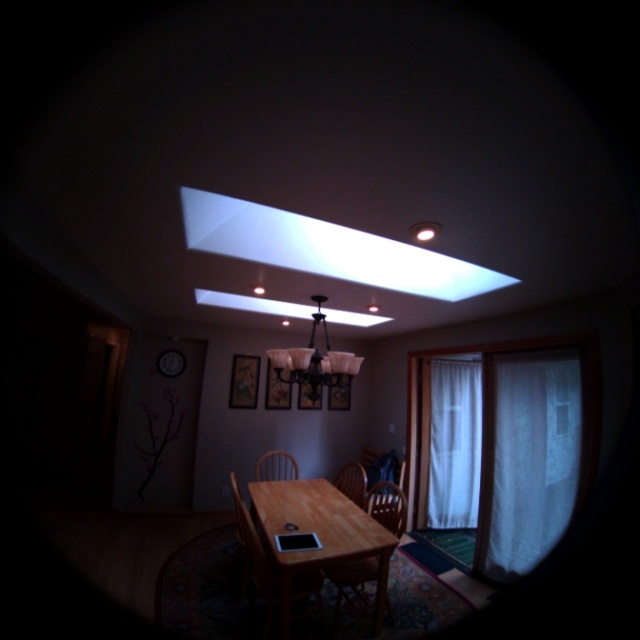}};
    \node(p12) at (p11.east) [anchor=west]{\includegraphics[width=\iw]{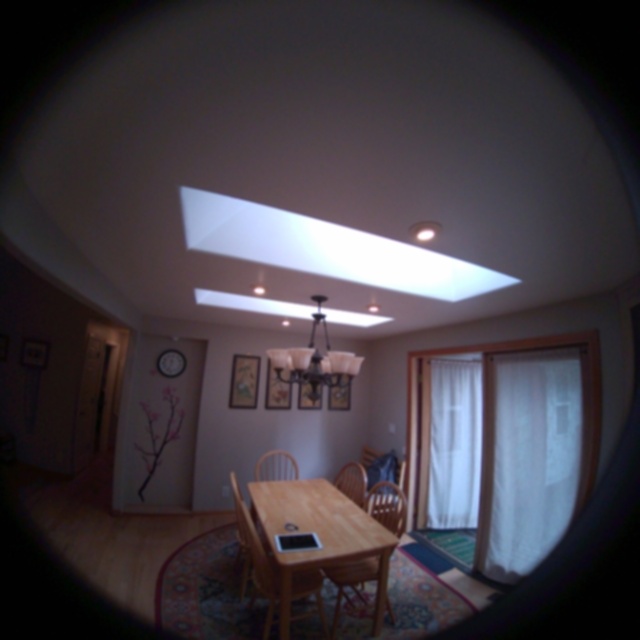}};
    \node(p13) at (p12.east) [anchor=west]{\includegraphics[width=\iw]{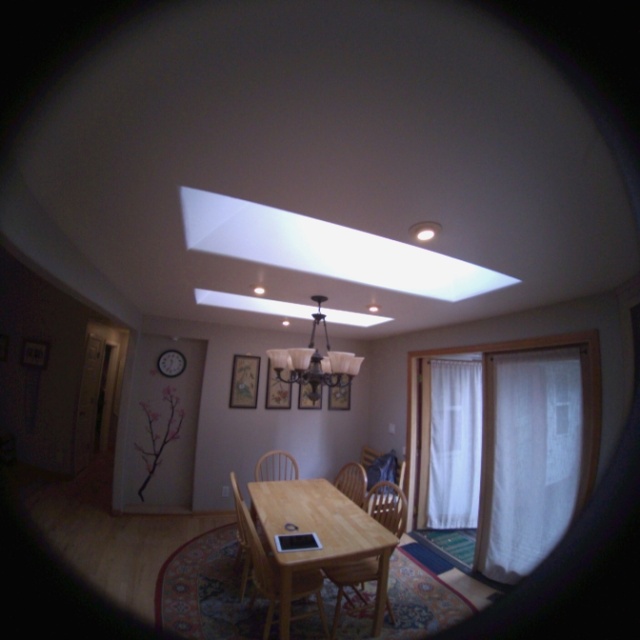}};
    \node(p14) at (p13.east) [anchor=west]{\includegraphics[width=\iw]{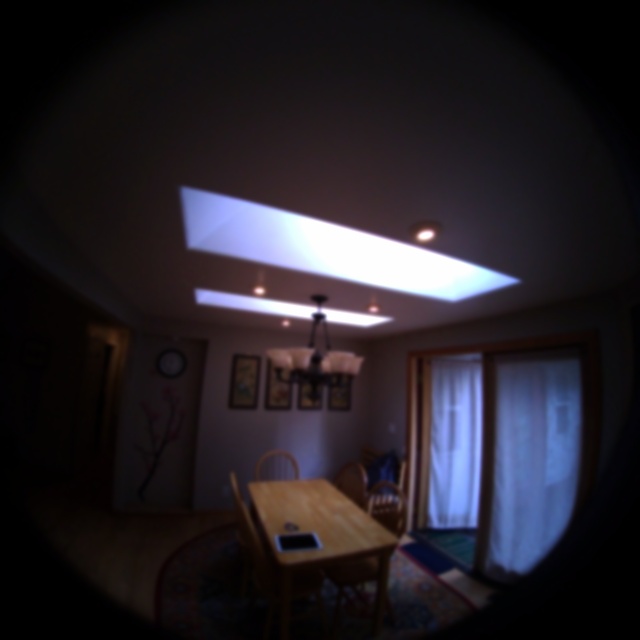}};

    % Row 2: Camera
    \node(c11) at (p11.south) [anchor=north,yshift=\dy]{\includegraphics[width=\iw]{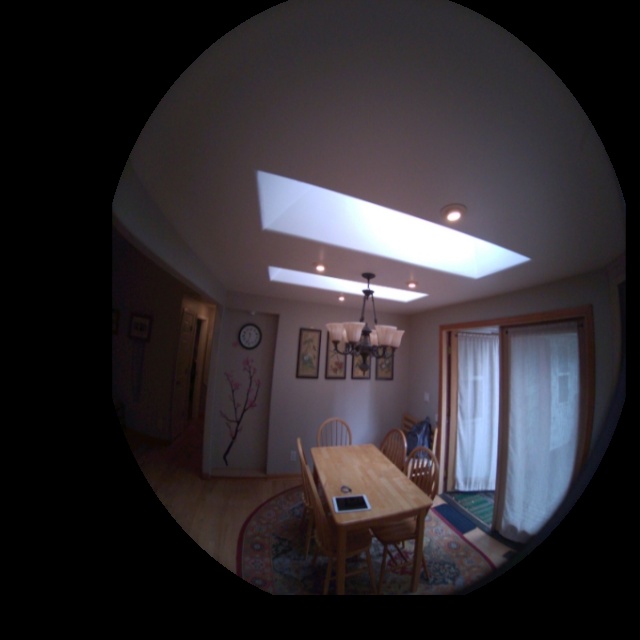}};
    \node(c12) at (p12.south) [anchor=north,yshift=\dy]{\includegraphics[width=\iw]{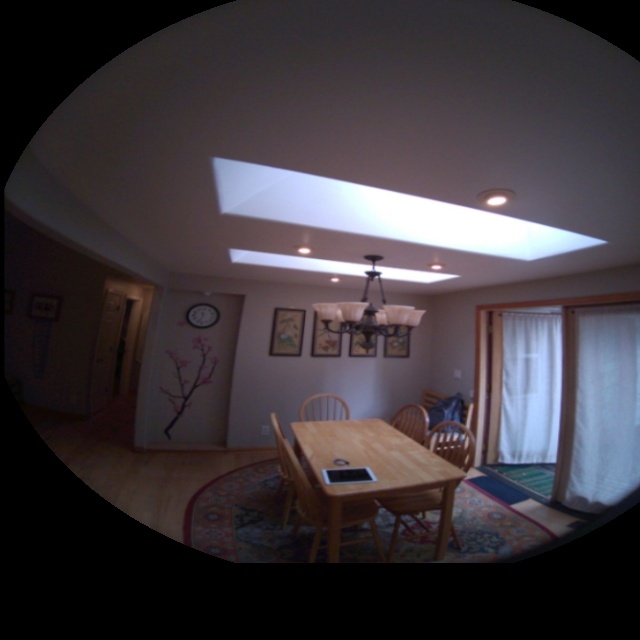}};
    \node(c13) at (p13.south) [anchor=north,yshift=\dy]{\includegraphics[width=\iw]{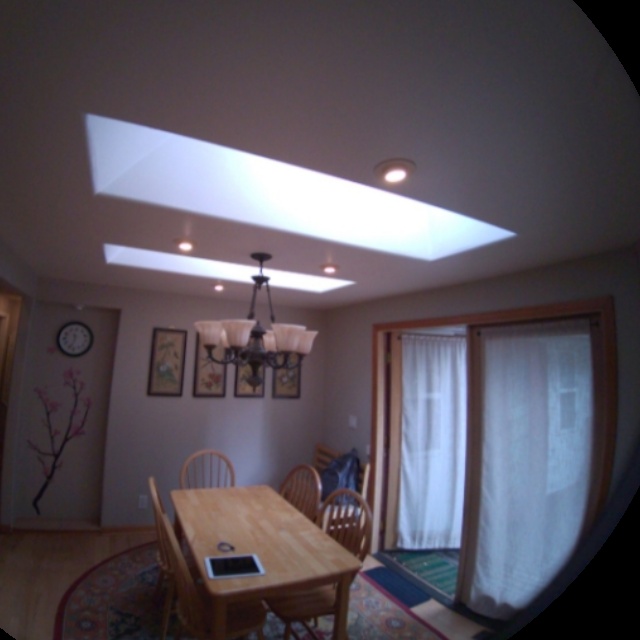}};
    \node(c14) at (p14.south) [anchor=north,yshift=\dy]{\includegraphics[width=\iw]{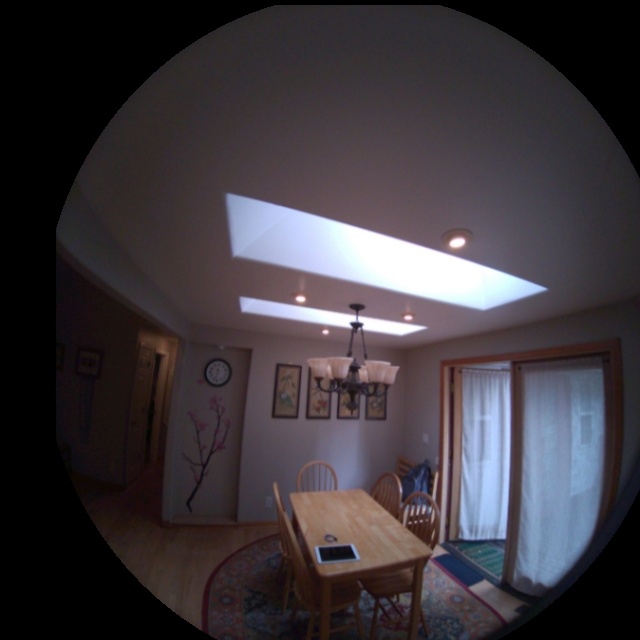}};

    % Row 3: 3D Point
    \node(d11) at (c11.south) [anchor=north,yshift=\dy]{\includegraphics[width=\iw]{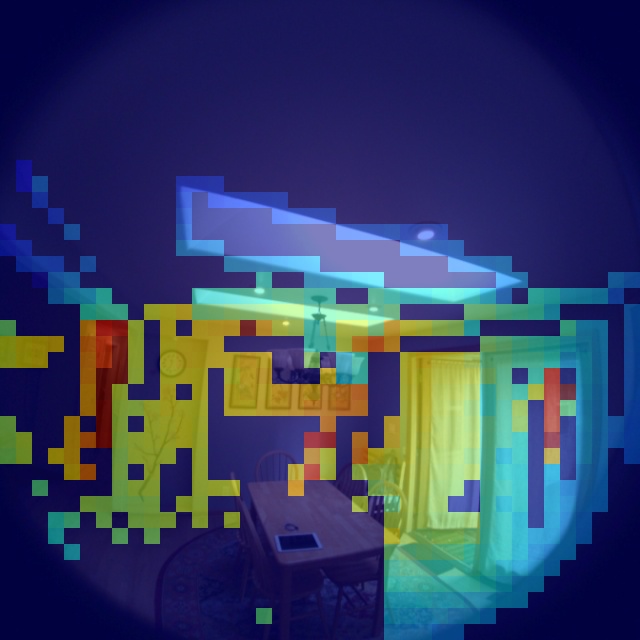}};
    \node(d12) at (c12.south) [anchor=north,yshift=\dy]{\includegraphics[width=\iw]{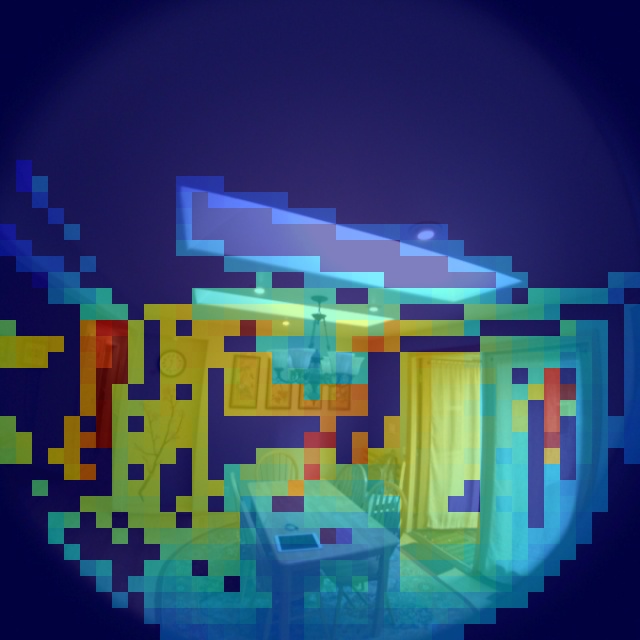}};
    \node(d13) at (c13.south) [anchor=north,yshift=\dy]{\includegraphics[width=\iw]{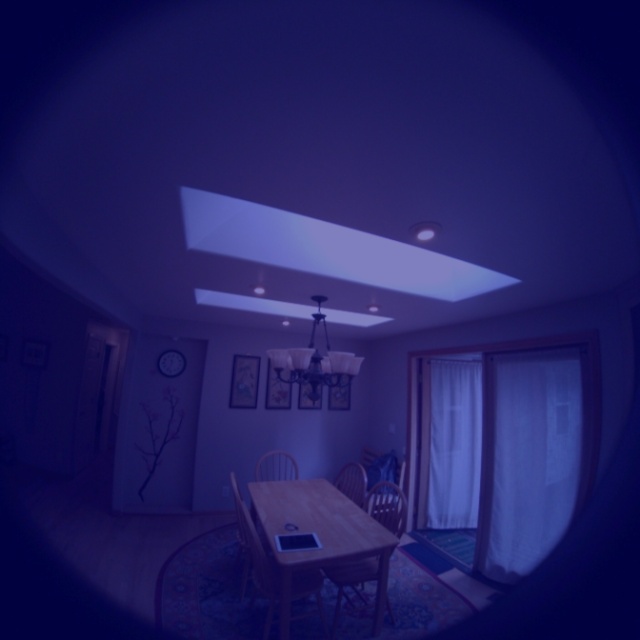}};
    \node(d14) at (c14.south) [anchor=north,yshift=\dy]{\includegraphics[width=\iw]{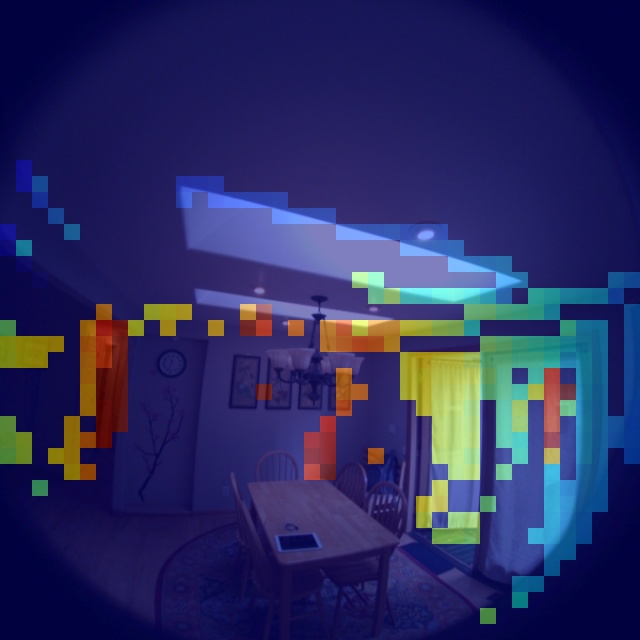}};

    % Row 4: 2D Box
    \node(b11) at (d11.south) [anchor=north,yshift=\dy]{\includegraphics[width=\iw]{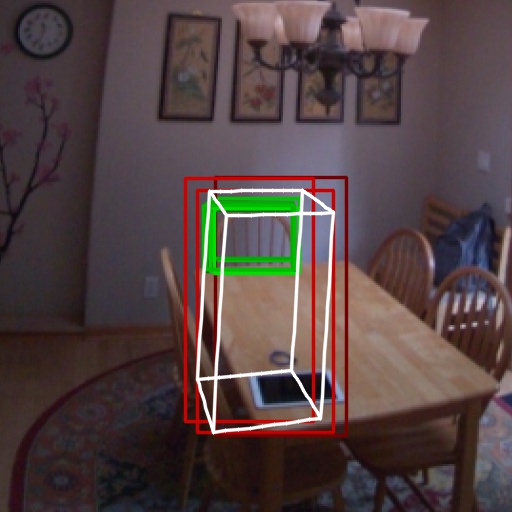}};
    \node(b12) at (d12.south) [anchor=north,yshift=\dy]{\includegraphics[width=\iw]{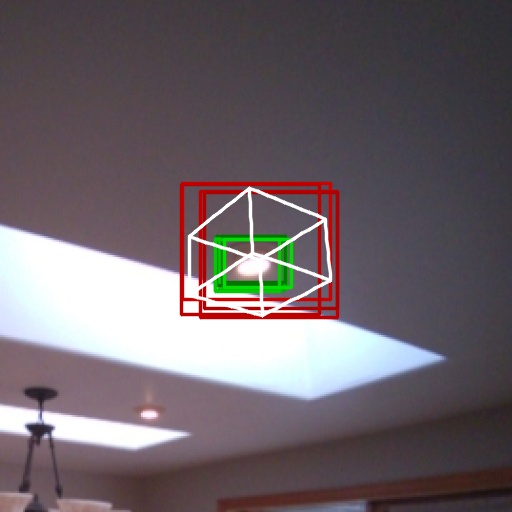}};
    \node(b13) at (d13.south) [anchor=north,yshift=\dy]{\includegraphics[width=\iw]{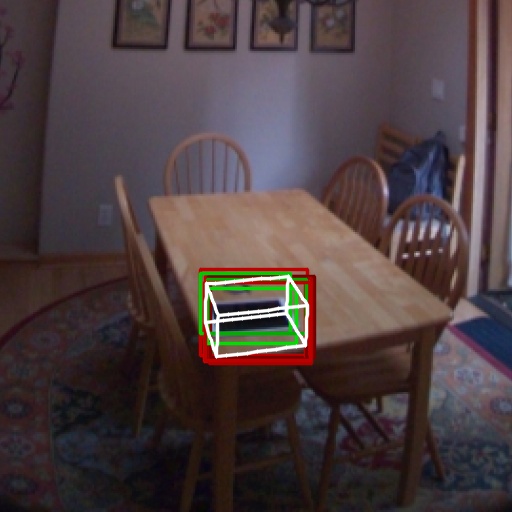}};
    \node(b14) at (d14.south) [anchor=north,yshift=\dy]{\includegraphics[width=\iw]{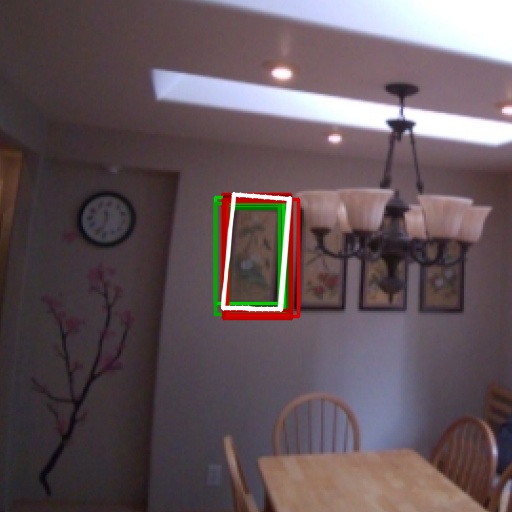}};

    % Column labels
    \node(t1) at (p11.north) [anchor=south, text width=\iw, align=center, font=\tiny]{Example 1};
    \node(t2) at (p12.north) [anchor=south, text width=\iw, align=center, font=\tiny]{Example 2};
    \node(t3) at (p13.north) [anchor=south, text width=\iw, align=center, font=\tiny]{Example 3};
    \node(t4) at (p14.north) [anchor=south, text width=\iw, align=center, font=\tiny]{Example 4};

    % Row labels
    \node(l1) at (p11.west) [anchor=south, rotate=90, yshift=-\dy, font=\tiny]{Photometric};
    \node(l2) at (c11.west) [anchor=south, rotate=90, yshift=-\dy, font=\tiny]{Camera};
    \node(l3) at (d11.west) [anchor=south, rotate=90, yshift=-\dy, font=\tiny]{3D Point};
    \node(l4) at (b11.west) [anchor=south, rotate=90, yshift=-\dy, font=\tiny]{2D Box};
  \end{tikzpicture}
  \caption{\textbf{Data augmentation examples.} Four augmentation types are shown by row (Photometric, Camera, 3D Point, and 2D Box), with four different examples per type shown by column.}
  \label{fig:aug-grid}
\end{figure}
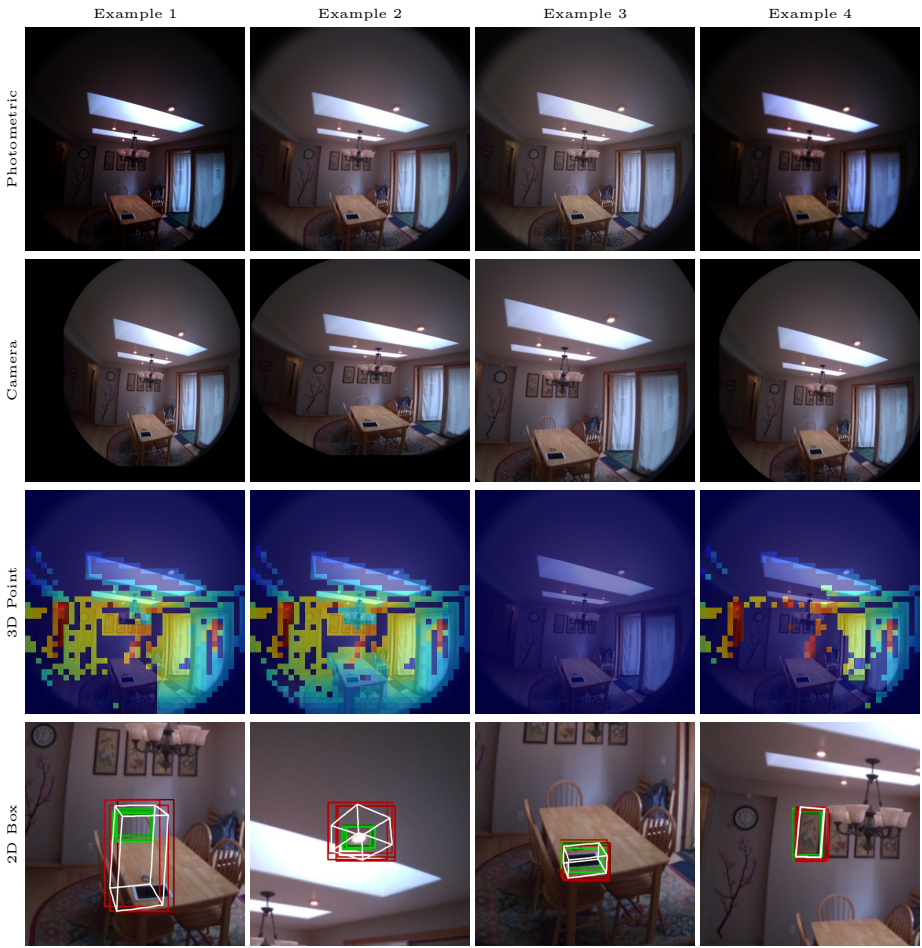

% \js{a figure showing the kinds of samples we get would speak volumes. especially on the camera calib augmentation}
First, we apply standard photometric augmentations to the input image, including random variations
in brightness, contrast, sharpness, Gaussian blur, and gamma, as shown in row 1 of ~\cref{fig:aug-grid}.

Second, we apply camera intrinsics augmentation by randomly perturbing the camera center and focal length, encouraging invariance to calibration noise and minor camera parameter errors. This supports operation on fisheye and non-fisheye distorted images. Various examples of this are shown in the second row of ~\cref{fig:aug-grid}.

Third, we augment the projected 3D points / dense depth by randomly dropping all points, individual points,
or contiguous blocks of points in 3D space, simulating sparsity and partial observability in the
geometric input. We visualize this in the third row of ~\cref{fig:aug-grid} by showing the depth patches (colored in a jet colormap). Note the third column has all points dropped out (to help encourage better RGB-only performance), while the fourth column has a large portion of the 3D dropped out, to support dynamics or scene content with no point cloud coverage to encourage robustness.

% \js{This builds a data "bridge" between the pinhole and fish-eye camera datasets. Do you have ablation without these augmentations?}
%
Lastly, we apply a novel type of data augmentation to account for the fact that some dataset do not have annotated, tight 2D bounding boxes.  A naive approach to generating 2DBBs projects the eight corners of the 3D box and takes the enclosing 2D bounding box; however, this often produces loose boxes and fails to account for occlusions. To address this, we prompt a segmentation model (SAM~\cite{kirillov2023segment}) using the projected 2D bounding boxes and derive tighter 2D bounding boxes from the resulting masks. We visualize examples of the 2D Box augmentation in \cref{fig:aug-grid} in the bottom most row. We visualize the original 3D box (white) and apply random Gaussian noise to a handful of examples to both of these non-tightened boxes (\textcolor{red}{red}) and SAM-tightened boxes (\textcolor{green}{green}). Different objects are highlighted in different rows.
In the first (leftmost) column, the chair is heavily occluded by the table, so there is a large difference in the SAM-tightened vs non-tightened boxes. In the second column, there are no occlusions, but the 3DBB annotation around the light is relatively loose, thus gets tightened by SAM. In the third and fourth columns, the SAM-tightened and non-tightened boxes are relatively similar, so SAM tightening does not have much effect.
Overall the purpose of this SAM-tightening procedure is to produce more realistic supervision that better matches detector outputs at inference time.
% \js{mention that you only keep high confidence 2DBBs}
%
% An example is shown in \cref{fig:aug-grid} in the last row. We apply this at random during training to allow the model to work with both tightened and non-tightened 2DBB prompts. We additionally apply Gaussian noise to the four corners of the resulting 2D bounding boxes input to the decoder during training time to further improve robustness to localization errors.

% \begin{figure}[!htbp]
% \centering
% \includegraphics[width=0.7\textwidth]{figures/misc/sam2daug.jpg}
% \caption{\textbf{SAM 2DBB Augmentation.}
% SAM is used to generate realistic tight 2DBBs from GT 3DBB annotations without the need for 2D manual annotation.\js{spruse up figure} \lingni{I'll remove the middle text and simply show more representative examples.} \js{I think remove the middle and just show a few 2DBB refinements: maybe top row is 2DBB from 3DBB and bottom row 2DBB refined via SAM? You could also show all 3 drawings in one frame if it oesnt get to cluttered and then just show a single row of examples} }
% \label{fig:sam2d}
% \end{figure}

% \subsection{Boxer Fisheye vs Pinhole on CA-1M} compare fisheye vs pinhole modes \TODO{computes this}.

% \subsection{Boxer Sparse vs Dense Depth on CA-1M} sparse vs dense depth modes \TODO{computes this}.

\section{Additional Experiment Details}

\subsection{Dataset Details}

\myparagraph{NymeriaPlus}

We use all basemaps and the participant Aria recordings from all non-basemap sequences, excluding all wrist-camera video. Of the five densely annotated locations, we use three for training and two for testing, i.e., location 10 and 44. For per-frame evaluation, we downsample the images to 1 Hz.

\myparagraph{Aria Digital Twin}. We use the sequence \nolinkurl{1WM103600M1292_optitrack_release_work_seq106} for testing. We exclude all dynamic objects (i.e., any object that is moved at any point during the recording) and evaluate only static objects. The ground-truth 2DBBs are generated only for static objects as well. To avoid unrealistic per-frame visibility, we only consider objects within 5 meters of camera depth. For per-frame evaluation, we temporally downsample the images from 30 Hz to 1 Hz.

% \lingni{only 1 sequence from ADT? Also fixed the text overflow, maybe move it into footnote.}
% dd -- yes only one ADT sequence, it is pretty long

\myparagraph{CA-1M}. We create a test set from the first 10 sequences in the validation split. The .tar file names used are: ca1m-val-45662921, ca1m-val-45261179, ca1m-val-47115543, ca1m-val-45261143, ca1m-val-45261615, ca1m-val-42897545, ca1m-val-45261133, ca1m-val-42897552, ca1m-val-45663113, and ca1m-val-42897521.

\subsection{Open-World Detector Settings}

OWLv2 is run using $960\times960$ resolution images, and DETIC is run using $800\times800$ input images. LVIS~\cite{gupta2019lvis} is used to generate a list of ~1200 commonly found objects to prompt OWLv2, excluding large structural elements like walls, floor, ceilings (see supp. for full list). We run CuTR \cite{lazarow2024cubify} and EVL \cite{straub2024efm3d} with their default settings. To create the LVIS+ prompt list, we append the following categories to LVIS: overhead\_light, recessed\_light, window, door, washer\_dryer, stairs, storage, shelf, plant, tree, electrical\_outlet, smoke\_detector, light\_switch, screen, tv, display, and smart\_phone.

\subsection{Baseline Method Settings}

\myparagraph{CuTR}. By default, CuTR detects its own 2DBBs using a DETR-style detection head. We found this detector slightly overfit to CA-1M, and performance improved when we overrode its 2DBBs with an off-the-shelf detector such as OWLv2 on non-CA1M data. Images are pinhole-rectified with focal length 850 and resized to the default CA-1M size (720 $\times$ 1024) to best match the CA-1M format. CuTR per-frame detections are fused into 3D using the same fusion system described in ~\cref{sec:offline}.

% \TODO{justify that vainCUTR/}

\myparagraph{EVL}. For EVL, we run using the default settings with 1 sec video snippets and multiple views for timestamp if available.

\myparagraph{3D-MOOD, Cube-RCNN}. We use the default settings to run 3D-MOOD and Cube-RCNN. We rectify all non-pinhole data to remove fisheye distortion before running.

\section{Additional Experiments}

\subsection{Example Open-World Pseudo Annotation to ScanNet}
A promising capability of a generic open-world 3DBB detector such as Boxer (trained on many camera types) is to run it on existing closed-world datasets and augment them with denser pseudo-annotations, which could support training larger models such as VLMs in the future.

ScanNet provides closed-world annotations via the Scan2CAD (oriented) 3DBB \cite{avocado2019scan2cad} annotations. These annotations cover the typical common classes (chair, bed, table, etc) but do not cover open world objects.

\begin{figure}[!htbp]
  \centering
  \def\iw{18em}
  \def\dy{-1pt}
  \begin{tikzpicture}[inner sep=1pt]
    % Row 1
    \node(p11)                            {\includegraphics[width=\iw]{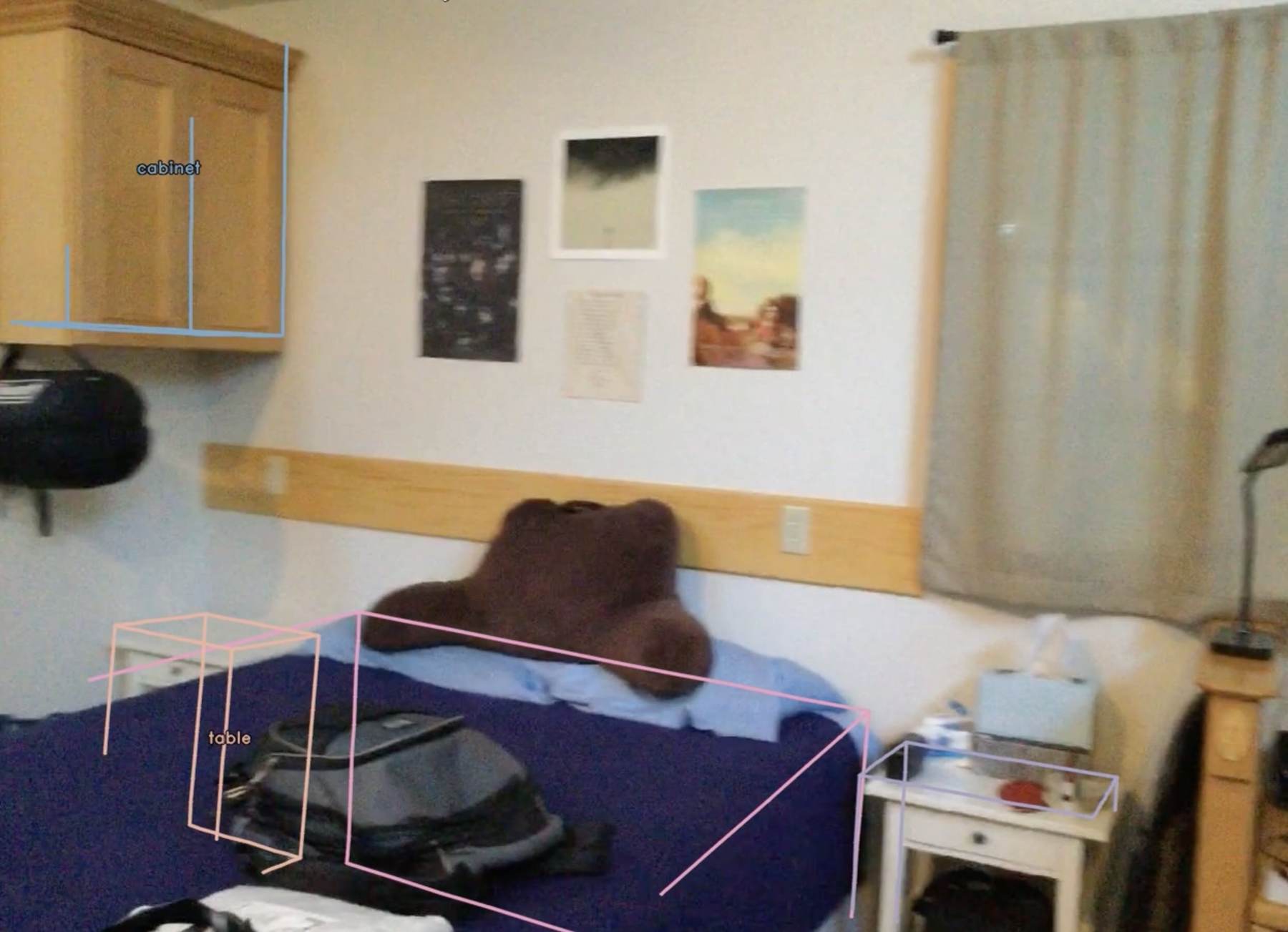}};
    \node(p12) at (p11.east) [anchor=west]{\includegraphics[width=\iw]{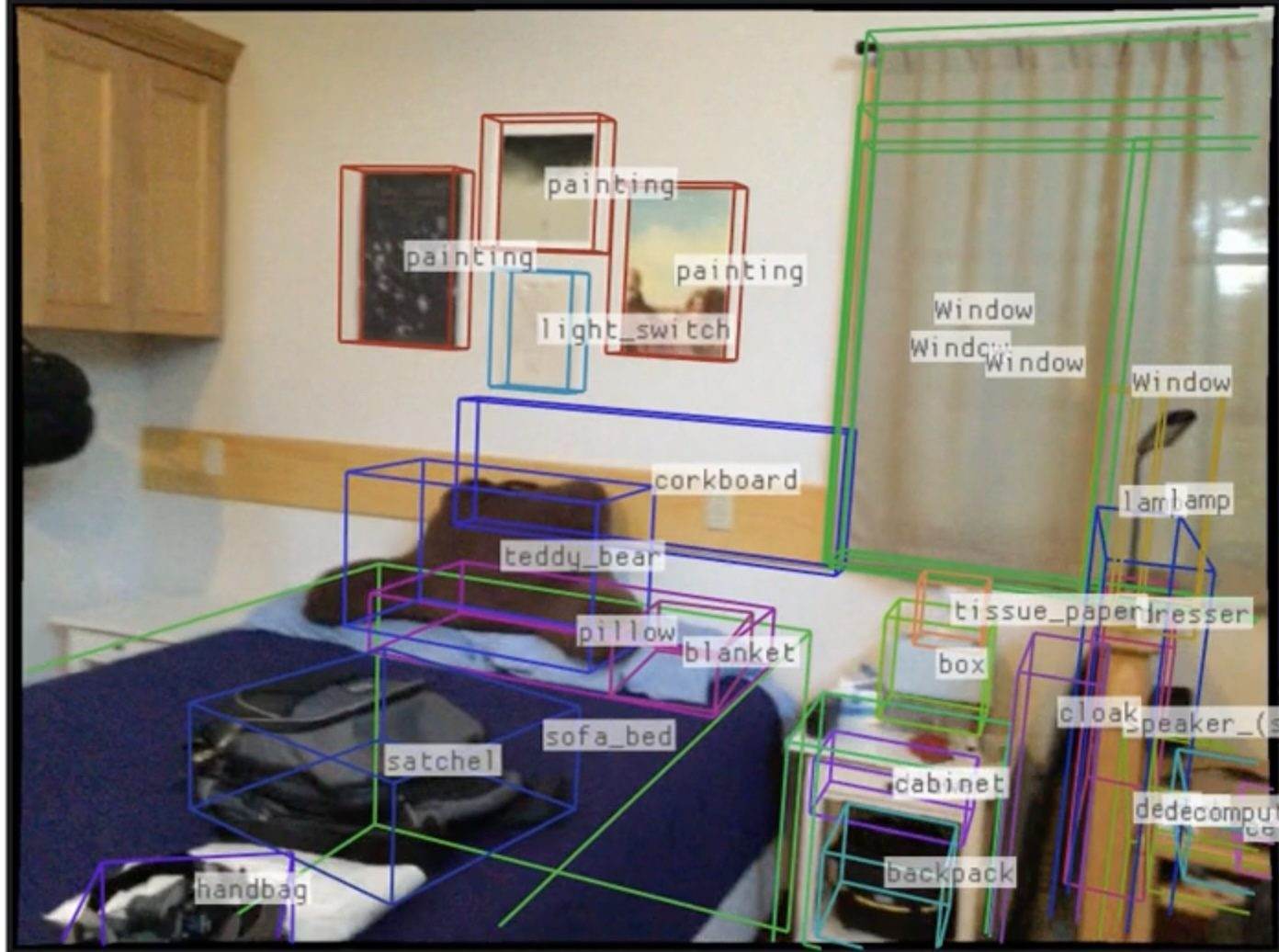}};

    % Column labels
    \node(t1) at (p11.north) [anchor=south, text width=\iw, yshift=-\dy, align=center, font=\tiny]{Closed-Set Existing Annotations};
    \node(t2) at (p12.north) [anchor=south, text width=\iw, yshift=-\dy, align=center, font=\tiny]{Open-Set Pseudo Annotations (ours)};

    % Row label
    \node(l1) at (p11.west) [anchor=south, rotate=90, yshift=-\dy, font=\tiny]{Example};
  \end{tikzpicture}
  \caption{\textbf{Pseudo open-set annotation examples on ScanNet.} We compare existing closed-set ScanNet annotations (left column) with pseudo open-set ScanNet annotations generated by Boxer (right column). One representative example is shown.}
  \label{fig:open-scannet}
\end{figure}
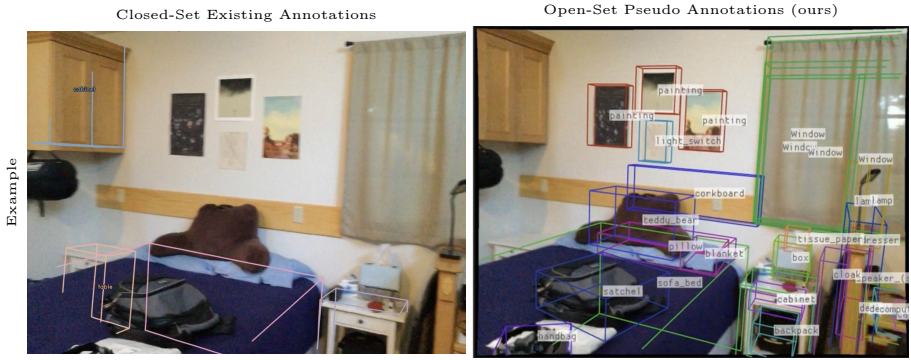

We show a few examples of adding such pseudo-annotations to ScanNet in ~\cref{fig:open-scannet}. Note that we focus primarily on getting the geometry correct. The semantic labels come from averaging OWLv2 predictions. These semantic categories could be further improved using a modern VLM using multi-view aggregation techniques such as \cite{luo2023scalable3dcaptioning}.

\subsection{Comparison to SAM3+SAM3D}
\label{sec:sam3d}
In this section, we compare BoxerNet against SAM3D for the 3DBB lifting task (termed layout estimation in \cite{chen2025sam}) using SAM3 to generate open-vocabulary object detections and segmentation masks. We report these results separately because both models are relatively slow (e.g., SAM3 takes about 45 s per image with 1000+ prompts, and SAM3D takes about 25 s per detected image). Therefore, we evaluate on a subset of 100 images from Aria Digital Twin with a reduced taxonomy: "chair, table, window, lamp, picture frame, sofa, book, container, shelf, and TV." For a fair comparison, we use the RGB+Depth input configuration and replace SAM3D's estimated point map with the dataset's dense ground-truth depth map.

\begin{figure}[!htbp]
  \centering
  \setlength{\fboxsep}{0pt}\fbox{\includegraphics[width=0.84\textwidth]{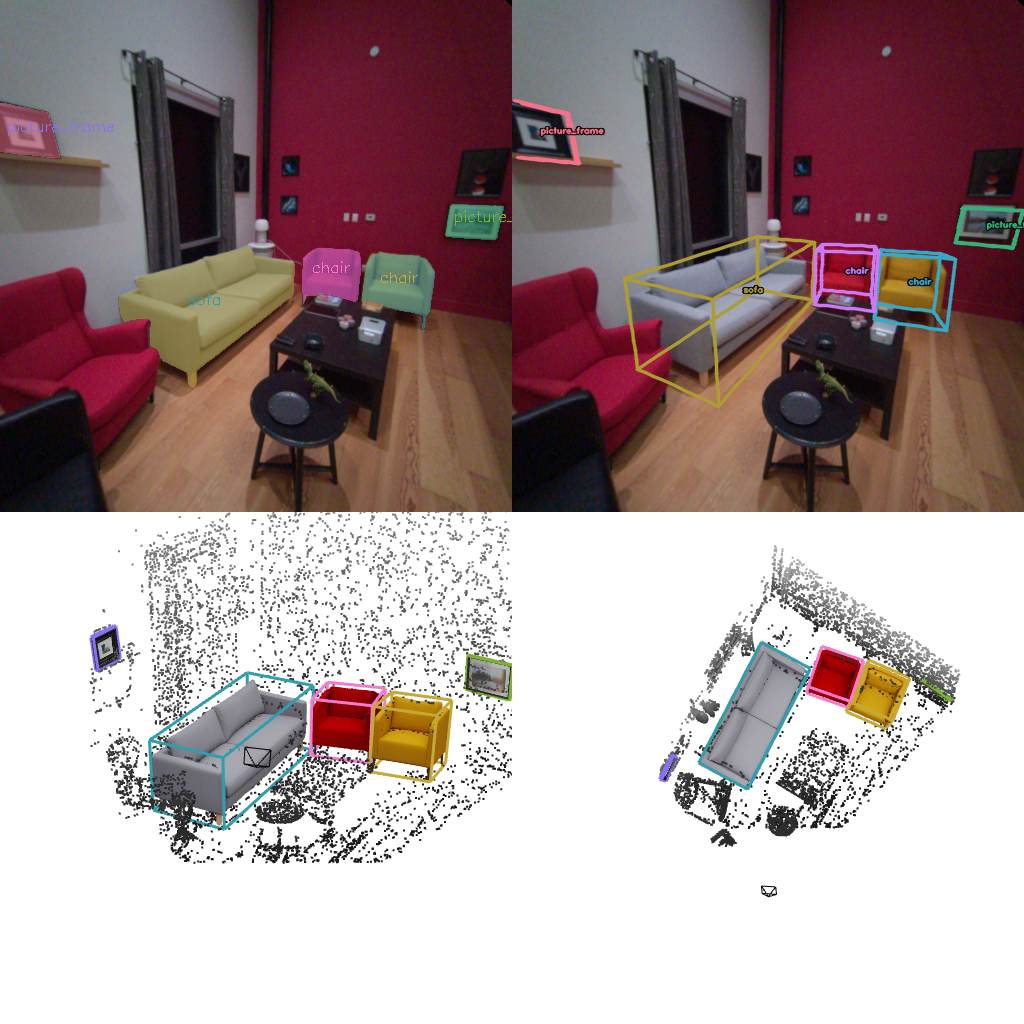}}
  \caption{\textbf{Example visualization of SAM3D on ADT.}
    Example output of SAM3+SAM3D (RGB+Depth) on Aria Digital Twin. Top-left shows SAM3 masks; top-right shows projected 3DBBs; bottom row shows two 3D views of predictions overlaid on the point cloud (bottom-left: behind view, bottom-right: bird's-eye view).
  }
  \label{fig:sam3d_example}
\end{figure}

\begin{table}[!htbp]
  \centering
  \scriptsize
  \setlength{\tabcolsep}{8pt}
  \renewcommand{\arraystretch}{1.05}
  \caption{\textbf{BoxerNet vs SAM3D.} We compare BoxerNet vs SAM3D class-agnostic per-frame mAP and precision (P) and recall (R) @ IoU=0.2 on a subset of 100 ADT images with a reduced set of text prompts.}
  \begin{tabular}{l c c c}
    \toprule
    Method & mAP & P@0.2 & R@0.2 \\
    \midrule
    SAM3+SAM3D (RGB+Depth)    & 0.027 & 0.540 & 0.058 \\
    SAM3+BoxerNet (RGB+Depth) & \textbf{0.064} & \textbf{0.822} & \textbf{0.088} \\
    \bottomrule
  \end{tabular}
  \label{tab:sam3d_adt}
\end{table}

Both methods use the same SAM3 detections as input. Because the taxonomy is limited while evaluation includes all objects, recall is expectedly low. Overall, BoxerNet achieves more accurate layout estimation, improving mAP from 0.027 to 0.064.

\subsection{mAP by Object Size}

For the class-agnostic evaluation above, all objects are grouped into one class, which can obscure performance differences across object scales. To analyze this, we partition CA-1M ground-truth boxes by volume into three buckets: small ($V<0.01\,\text{m}^3$), medium ($0.01\leq V\leq0.1\,\text{m}^3$), and large ($V>0.1\,\text{m}^3$). Interpreted as equivalent cube side lengths ($s=\sqrt[3]{V}$), these thresholds correspond to approximately $s<0.215$ m (small), $0.215\leq s\leq0.464$ m (medium), and $s>0.464$ m (large).

\begin{figure}[!htbp]
  \centering
  \includegraphics[width=1.0\textwidth]{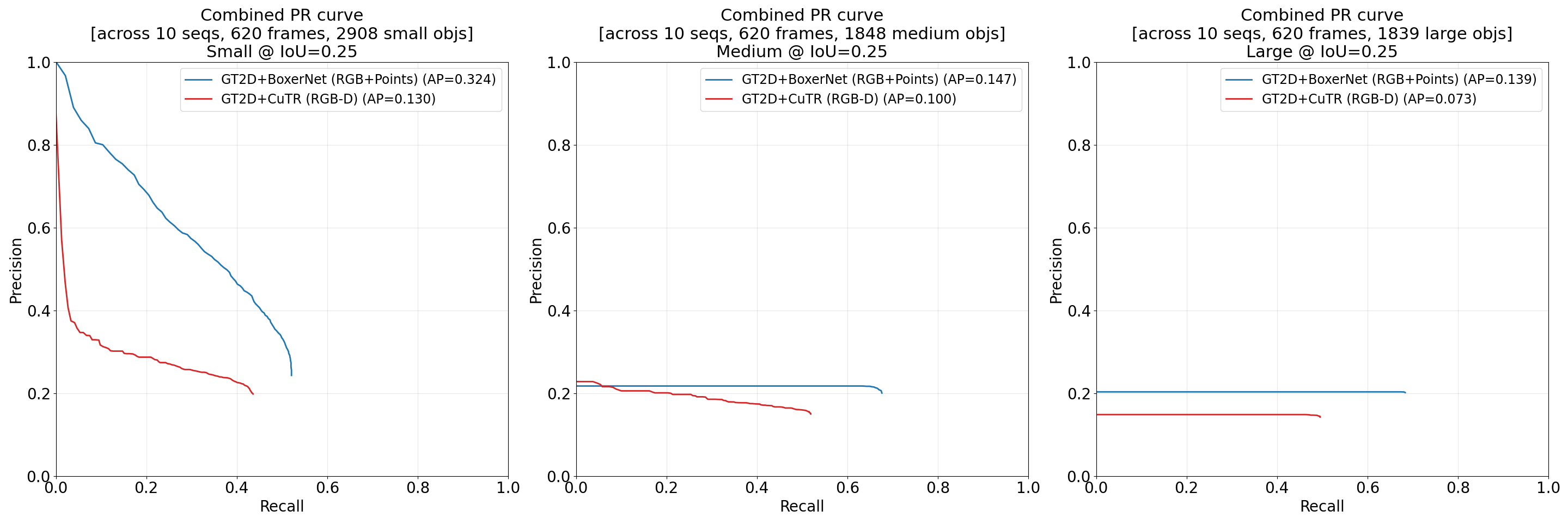}
  \caption{\textbf{PR Curve By Object Size.}
  Precision--recall curves for three object-size buckets---small (left), medium (middle), and large (right)---for BoxerNet vs. CuTR. }
  \label{fig:pr_by_size}
\end{figure}

Based on the PR curves in ~\cref{fig:pr_by_size}, small objects account for most of the performance gap between BoxerNet and CuTR at an IoU threshold of 0.25. For medium and large objects, the relatively flat curves indicate that each method tends to either detect an object correctly or miss it entirely. Small objects are also the most common, with 2908 small, 1848 medium, and 1839 large objects.

\section{Acknowledgments}
% \lingni{make sure you hide this section from supp. Also double check the meta data of the video and remove any personal information to prevent desk rejection. }
We thank Pierre Moulon for feedback in group discussions, Manuel Lopez Antequera for support with internal annotations, Dan Barnes and Raul Mur-Artal for support with tooling for annotation, and Yawar Siddiqui for valuable discussions and feedback.  We also thank Austin Kukay, Rowan Postyeni, Ruosha Pang, Mu Cheng, William Sun, Chen
Zhang, Qinyue He, Aaron Deguzman, Yao Zhi, Luis Pesqueira, Abha Arora, and Rana Hayek for annotation
support.

\clearpage
\bibliographystyle{splncs04}
\bibliography{main}
\end{document}